\RequirePackage{silence}
\PassOptionsToPackage{svgnames,dvipsnames,table}{xcolor}
\PassOptionsToPackage{capitalize}{cleveref}
\documentclass[table]{ai2style/ai2}
\usepackage{amssymb}
\usepackage{bigdelim}
\usepackage{todonotes}
\usepackage{longtable}
\usepackage{float}
\usepackage{tabularx}
\usepackage{tabularray}
\usepackage[most]{tcolorbox} 
\usepackage{svg}
\usepackage[absolute]{textpos}
\usepackage{fdsymbol}
\usepackage[utf8]{inputenc}
\usepackage[T1]{fontenc} 
\usepackage{url}
\usepackage{booktabs}    
\usepackage{amsfonts}
\usepackage{nicefrac}    
\usepackage{microtype}   
\usepackage[table]{xcolor}
\usepackage{amsmath}
\DeclareMathSizes{11}{11}{8}{6}
\usepackage{csquotes}

\usepackage{siunitx}
\usepackage{graphicx}
\usepackage{arydshln}
\usepackage{wrapfig}
\usepackage{enumitem}
\usepackage{soul} %

\usepackage{multirow}
\usepackage{xspace}
\usepackage{adjustbox}
\usepackage{pifont}
\usepackage{makecell}
\usepackage{bold-extra}

\usepackage{caption}

\usepackage{hyperref}
\definecolor{linkcolor}{RGB}{0, 0, 128}
\hypersetup{
     colorlinks   = true,
     citecolor    = linkcolor,
     linkcolor    = linkcolor,
     urlcolor     = linkcolor,
}
\usepackage{listings}

\setlist[itemize]{leftmargin=*,itemsep=0em,parsep=0.3em,topsep=0.3em}

\definecolor{iclrdeepblue}{RGB}{0,0,128}
\definecolor{maroon}{HTML}{F26035}
\definecolor{yellow}{HTML}{FDBC42}
\definecolor{lavender}{HTML}{734f96}
\definecolor{darkergrey}{HTML}{444444}
\definecolor{midgrey}{HTML}{e6eded}
\definecolor{ai2pink}{HTML}{5477C4}%
\definecolor{ai2midpink}{HTML}{fad3e5}
\definecolor{ai2lightpink}{HTML}{fbecf3}
\definecolor{ai2midwhite}{HTML}{f2e5d9}
\definecolor{ai2offwhite}{HTML}{fbf4ee}
\definecolor{ai2green}{HTML}{0fcb8c}
\definecolor{ai2lightgreen}{HTML}{e7f9f3}
\definecolor{ai2darkgreen}{HTML}{105257}
\definecolor{ai2purple}{HTML}{B932EB}
\definecolor{ai2lightpurple}{HTML}{f7e8fc}
\definecolor{neutralEight}{HTML}{343434}
\definecolor{neutralFive}{HTML}{838383}
\definecolor{neutralThree}{HTML}{bebebe}
\definecolor{neutralOne}{HTML}{dedede}
\definecolor{lightgrey}{HTML}{fafcfc}

\usepackage{tikz}
\usepackage[framemethod=TikZ]{mdframed}
\usepackage{xpatch}
\usetikzlibrary{shadows}
\usepackage{pgfplots}
\pgfplotsset{compat=1.18}

\mdfdefinestyle{mdpurplebox}{%
  roundcorner=10pt,
  linewidth=1pt,
  skipabove=12pt,
  skipbelow=12pt,
  innertopmargin=9pt,
  innerbottommargin=9pt,
  linecolor=black,
  nobreak=true,
  backgroundcolor=Orchid!10,
  shadow=true,
  shadowsize=6pt,
  shadowcolor=black!30,
  frametitleaboveskip=8pt,
  frametitlebelowskip=8pt,
  frametitlebackgroundcolor=Violet!50!black,
  frametitlefont=\bfseries\sffamily\color{white},
  frametitlerule=true,
}
\makeatletter
\xpatchcmd{\endmdframed}
  {\aftergroup\endmdf@trivlist\color@endgroup}
  {\endmdf@trivlist\color@endgroup\@doendpe}
  {}{}
\makeatother

\DeclareUnicodeCharacter{2011}{\nobreakdash-}
\DeclareUnicodeCharacter{2013}{--}
\DeclareUnicodeCharacter{2019}{\textquoteright}
\DeclareUnicodeCharacter{201C}{``}
\DeclareUnicodeCharacter{201D}{''}
\DeclareUnicodeCharacter{202F}{\nobreak\hspace{0.16667em}}
\DeclareUnicodeCharacter{21D2}{\ensuremath{\Rightarrow}}
\DeclareUnicodeCharacter{220E}{\ensuremath{\square}}

\definecolor{darkred}{RGB}{156, 39, 33}
\definecolor{darkblue}{RGB}{31, 90, 153}
\definecolor{forestgreen}{rgb}{0.13, 0.55, 0.13}
\definecolor{olmoDarkBlue}{HTML}{012e59}
\definecolor{olmoBlue}{HTML}{265ed4}
\definecolor{olmoLightBlue}{HTML}{012e59}
\definecolor{olmoTeal}{HTML}{00d5ff}
\definecolor{olmoYellow}{HTML}{ffbb00}
\definecolor{olmoOrange}{HTML}{ff9100}

\newcommand{\huggingface}{\raisebox{-1.5pt}{\includegraphics[height=1.05em]{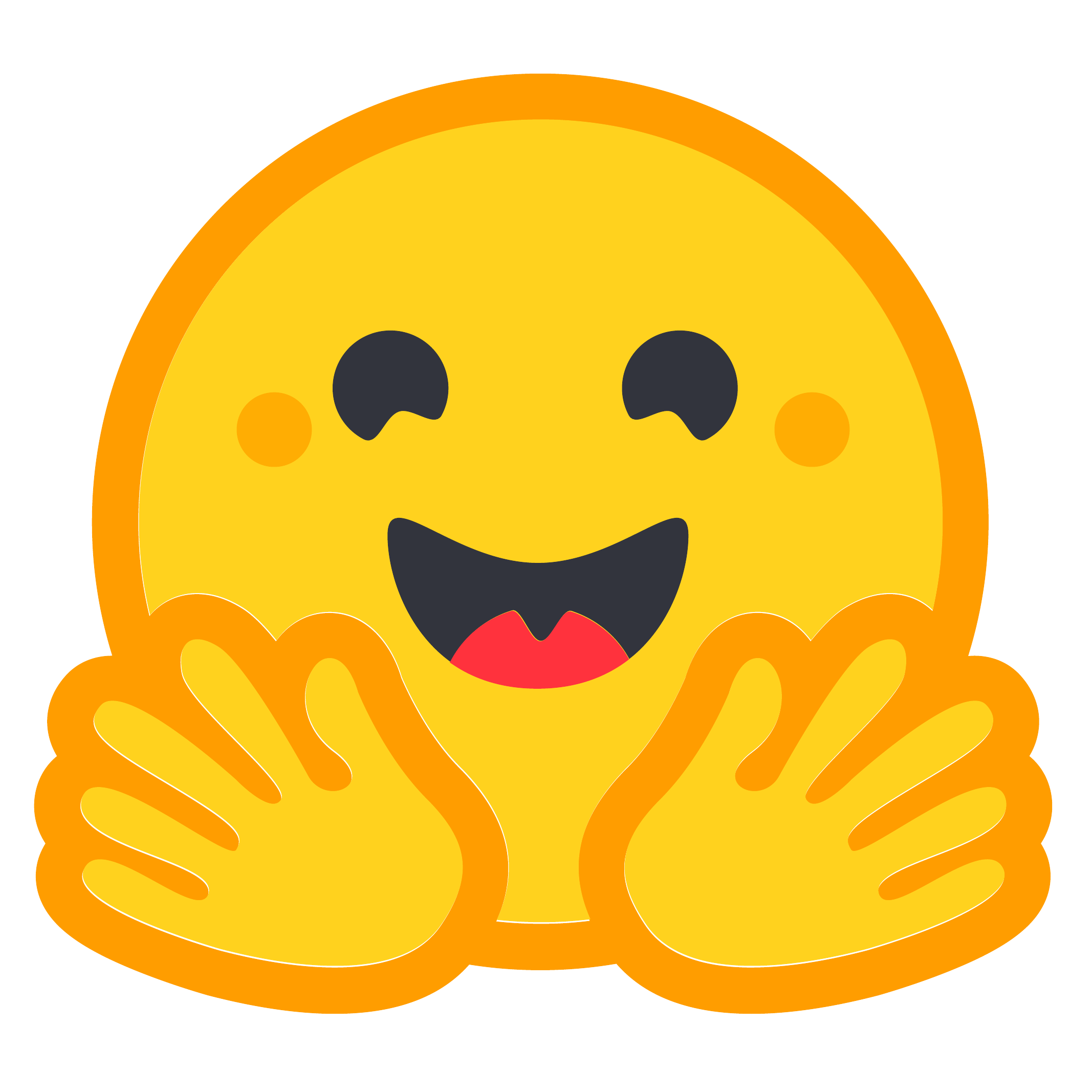}}\xspace}

\newcommand{\github}{\raisebox{-1.5pt}{\includegraphics[height=1.05em]{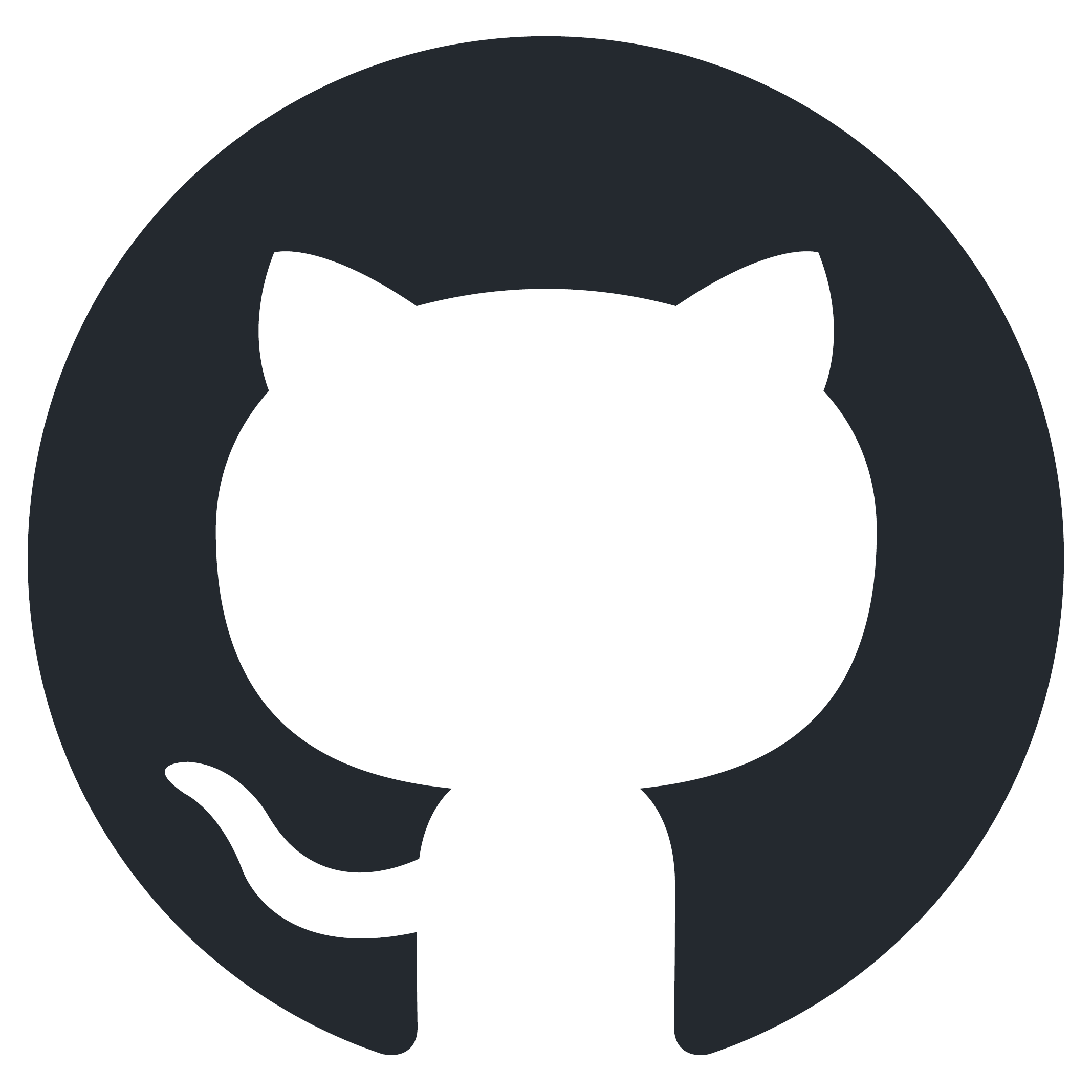}}\xspace}

\usepackage{setspace}

\usepackage{nicematrix}
\newcolumntype{L}[1]{>{\raggedright\let\newline\\\arraybackslash\hspace{0pt}}m{#1}}
\newcolumntype{C}[1]{>{\centering\let\newline\\\arraybackslash\hspace{0pt}}m{#1}}
\newcolumntype{R}[1]{>{\raggedleft\let\newline\\\arraybackslash\hspace{0pt}}m{#1}}
\newcolumntype{P}[1]{>{\centering\let\newline\\\arraybackslash\columncolor{ai2lightpink}}m{#1}}

\title{Loop the Loopies!}

\authorTwo{Zitian Gao}
\authorTwo{Yilong Chen}
\authorTwo{Yihao Xiao}
\authorTwo{Xinyu Yang}

\authorThree{Ran Tao}
\authorThree{Joey Zhou}
\authorThree{Bryan Dai}

\affiliation{\raisebox{2mm}[0pt][0pt]{IQuest Research}}

\metadata[\vspace{-8mm}]{\color{ai2accent}\hspace*{-1mm} See the full author contributions \hyperref[sec:contrib]{here}.}

\abstract{

We present Loopie, the most powerful open-source looped Transformer to date. The Loopie series consists of two Mixture-of-Experts (MoE) models: a 20B-parameter model with 2B active parameters and a 6B parameter model with 0.6B active parameters. Looped Transformers have long faced a challenge: given an $N$-fold pre-training compute, increasing the parameter count by a factor of $N$ usually outperforms looping $N$ times. Loopie addresses this challenge. Extensive ablation studies, including comparisons with a vanilla 30B-A3B model, show that Loopie substantially outperforms vanilla Transformer baselines trained with the same compute budget. With a novel post-training method, Loopie develops strong reasoning abilities and achieves frontier-level reasoning performance.

}

\metadata[\vspace{6mm}\quad\huggingface Models:]{
\href{https://huggingface.co/zitian-gao/Loopie-20B-A2B-preview}{\texttt{Loopie-20B-A2B}} \quad 
\href{https://huggingface.co/zitian-gao/Loopie-6B-A0.6B-preview}{\texttt{Loopie-6B-A0.6B}}}

\metadata[\vspace{.5em}\quad\github Code:]{
    \href{https://github.com/zitian-gao/megatron-lm}{\texttt{megatron-lm}} \quad
    \href{https://github.com/zitian-gao/transformers}{\texttt{transformers}} \quad
    \href{https://github.com/zitian-gao/vllm}{\texttt{vllm}}
}

\begin{document}
\maketitle

\vspace{6mm}

\begin{center}
    \hspace*{-2mm}\includegraphics[width=0.92\textwidth]{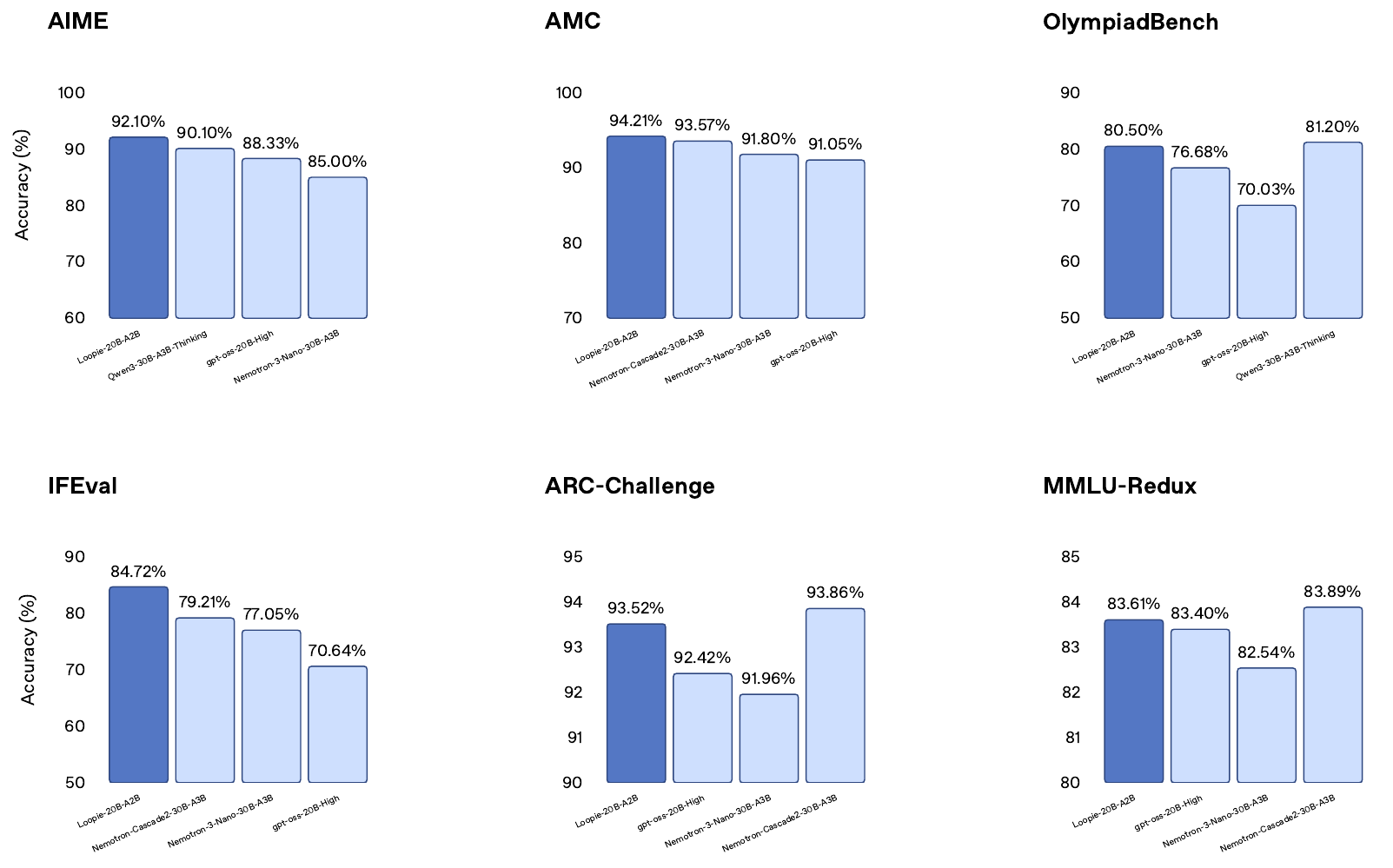}\hspace*{2mm}
\end{center}
\newpage
\setcounter{tocdepth}{2}
\tableofcontents
\newpage

\section{Introduction}
\label{sec:intro}

Looped Transformers, first introduced as Universal Transformers \citep{dehghani2018universal}, have recently re-emerged as a compelling alternative to conventional depth scaling. Rather than stacking distinct layers, they repeatedly apply the same model across recurrent steps, closely connecting this approach to parameter sharing in Transformers \citep{lan2019albert,dehghani2018universal}. This recurrent approach has shown strong empirical performance across a wide range of domains, including language modeling, algorithmic learning, and abstract reasoning \citep{gao2025universal,yang2023looped,jolicoeurmartineau2025trm,wang2025hierarchical,dehghani2018universal,saunshi2025latentthoughts,bay2025mixture,geiping2025scalinglatent,zhu2025scalinglatent,frey2026adaptive,huang2026equilibrium}.

A growing body of work suggests that recurrent computation is especially well suited to complex problems. Looped Transformers can outperform vanilla Transformers on in-context learning and data-fitting tasks, and they can implement multi-step gradient descent in context with far fewer parameters \citep{yang2023looped,fan2024looped,giannou2023looped,gatmiry2024can,gatmiry2024role,chen2025bypassing}. Studies of looped Transformers have also shown that recurrent computation allows shallow parameterizations to approach the performance of deeper untied models on reasoning tasks \citep{saunshi2025latentthoughts,jeddi2026loopformer,geiping2025scalinglatent,zhu2025scalinglatent}. These advantages are particularly visible on abstract reasoning tasks such as ARC-AGI, Sudoku, and Maze, where results from recent recurrent models suggest that recurrence provides a powerful inductive bias for compositional generalization, rule discovery, and inductive reasoning \citep{wang2025hierarchical,jolicoeurmartineau2025trm,gao2025universal}.

Recent work has begun scaling looped Transformers to billion-parameter language models. Ouro models scale to 1.4B and 2.6B parameters with four recurrent steps, reporting strong parameter efficiency relative to larger dense baselines \citep{zhu2025scalinglatent}. Huginn scales to 3.5B parameters and uses 32 recurrent steps to trade parameters for latent computation \citep{geiping2025scalinglatent}. Other systems further study adaptive recurrence, efficient reasoning, memory--compute trade-offs, and scaling laws for stable looped language models \citep{bay2025mixture,jeddi2026loopformer,frey2026adaptive,prairie2026parcae}. However, these gains expose a fundamental compute-accounting issue: \textbf{looping a model $N$ times during pre-training also multiplies pre-training compute by $N$. Thus, looped Transformers should be compared not only against vanilla Transformers with the same parameter count but also against non-looped models trained under the same pre-training compute budget.} This view is consistent with language-model scaling and compute-optimal training analyses, which evaluate quality as a function of parameters, data, and total training FLOPs rather than parameter count alone \citep{kaplan2020scaling,hoffmann2022training,prairie2026parcae}. For example, Ouro-2.6B with 4 loops should be compared against a baseline model with a parameter count close to 2.6B × 4 = 10.4B; likewise, after accounting for pre-training compute, Huginn-3.5B with 32 loops should be compared against a much larger 112B-parameter baseline model. This motivates our central question:

\begin{tcolorbox}[colback=ai2offwhite,colframe=ai2pink,boxrule=0.5pt,arc=3pt,left=6pt,right=6pt,top=5pt,bottom=5pt]
\centering
\emph{Can looped Transformers match or exceed vanilla Transformers \\
under the same pre-training compute budget?}
\end{tcolorbox}

We answer this question with the \textbf{Loopie Series}: two looped MoE LLMs, Loopie-20B-A2B and Loopie-6B-A0.6B, each trained with two loop steps. Our key idea is the \textbf{Loopie Recipe}, a compute-matched scaling recipe that addresses the main concern for recurrent-depth scaling: under a fixed pre-training compute budget, vanilla parameter scaling can otherwise dominate looping \citep{hoffmann2022training,prairie2026parcae}. With a novel post-training method, Loopie develops strong reasoning abilities and achieves frontier-level reasoning performance.

Our contributions are threefold:
\begin{itemize}
    \item \textbf{Compute-matched scaling.} We introduce the Loopie Recipe to address the fixed-compute challenge for looped Transformers and validate it through extensive ablations.
    \item \textbf{Scalable looped MoE models.} We demonstrate that looped computation scales to large MoE language models by training Loopie-20B-A2B and Loopie-6B-A0.6B.
    \item \textbf{Large-scale post-training.} We scale Loopie through large-scale post-training, including a novel supervised pre-training stage, it yields strong reasoning abilities.
\end{itemize}

\newpage

\usetikzlibrary{arrows.meta,positioning,calc,shapes.geometric}
\definecolor{delta_color}{RGB}{242,243,193}
\definecolor{swa_color}{RGB}{252,224,225}
\definecolor{glu_color}{RGB}{194,232,247}
\definecolor{silu_color}{RGB}{203,231,207}
\definecolor{linear_color}{RGB}{220,223,240}
\definecolor{conv_color}{RGB}{252,224,225}
\definecolor{gray_bbox_color}{RGB}{243,243,244}
\begin{figure*}[!t]
\centering
\vspace{15pt}
\tikzset{
  model/.style={
    draw=black, very thick, fill=gray_bbox_color,
    minimum width=96pt, rounded corners=8pt
  },
  gdelta/.style={
    draw=black, very thick, fill=gray_bbox_color,
    minimum width=120pt, minimum height=190pt, rounded corners=10pt
  },
  swa/.style={
    draw=black, very thick, fill=swa_color!80,
    minimum width=78pt, minimum height=0.7cm, rounded corners=3pt
  },
  glu/.style={
    draw=black, very thick, fill=glu_color!80,
    minimum width=78pt, minimum height=0.7cm, rounded corners=3pt
  },
  conv/.style={
    draw=black, very thick, minimum width=50pt, minimum height=20pt,
    fill=conv_color!80, rounded corners=3pt
  },
  linear/.style={
    draw=black, very thick, trapezium,
    trapezium left angle=110, trapezium right angle=110,
    minimum width=50pt, minimum height=15pt, fill=linear_color!80
  },
  branch_node/.style={
    draw=black, very thick, minimum width=40pt, minimum height=15pt,
    fill=linear_color!80, rounded corners=3pt
  },
  silu/.style={
    draw=black, very thick, fill=silu_color,
    minimum width=35pt, minimum height=12pt, rounded corners=3pt
  },
  layerlink/.style={-latex, line width=0.6pt},
  modulelink/.style={-latex, very thick, densely dashed},
  normlink/.style={very thick},
  otimes/.style={
    draw=black, very thick, circle, minimum size=11pt,
    inner sep=0pt, outer sep=0pt,
    path picture={
      \draw (path picture bounding box.center) -- ++(0.25cm,0.25cm)
            (path picture bounding box.center) -- ++(-0.25cm,-0.25cm)
            (path picture bounding box.center) -- ++(-0.25cm,0.25cm)
            (path picture bounding box.center) -- ++(0.25cm,-0.25cm);
    }
  },
  gradlink/.style={very thick, grad_color, -{Latex[length=3mm]}},
  stopgrad/.style={
    fill=gray_bbox_color,
    draw=grad_color, very thick, circle, minimum size=10pt, inner sep=0pt,
    path picture={
      \draw[grad_color, very thick]
        (path picture bounding box.south west) -- (path picture bounding box.north east)
        (path picture bounding box.north west) -- (path picture bounding box.south east);
    }
  }
}

\definecolor{grad_color}{RGB}{220,70,70}

\scalebox{0.85}{
\centering
\resizebox{0.7\textwidth}{!}{

\begin{tikzpicture}

\node[model, minimum width=84pt, minimum height=76pt] (model1) at (0,0) {};
\node[below=6pt, align=center] at (model1.south) {layer 1};
\node[swa, minimum width=0.7cm, minimum height=52pt] (attn1) at ($(model1.center) + (-21pt, 0)$) {\rotatebox{90}{Attention}};
\node[glu, minimum width=0.7cm, minimum height=52pt] (moe1) at ($(model1.center) + (21pt, 0)$) {\rotatebox{90}{MoE}};

\node[model, minimum width=84pt, minimum height=76pt] (model2) at (123.1pt,0) {};
\node[below=6pt, align=center] at (model2.south) {layer 2};
\node[swa, minimum width=0.7cm, minimum height=52pt] (attn2) at ($(model2.center) + (-21pt, 0)$) {\rotatebox{90}{Attention}};
\node[glu, minimum width=0.7cm, minimum height=52pt] (moe2) at ($(model2.center) + (21pt, 0)$) {\rotatebox{90}{MoE}};

\node[model, minimum width=84pt, minimum height=76pt] (model3) at (246.2pt,0) {};
\node[below=6pt, align=center] at (model3.south) {layer 3};
\node[swa, minimum width=0.7cm, minimum height=52pt] (attn3) at ($(model3.center) + (-21pt, 0)$) {\rotatebox{90}{Attention}};
\node[glu, minimum width=0.7cm, minimum height=52pt] (moe3) at ($(model3.center) + (21pt, 0)$) {\rotatebox{90}{MoE}};

\node[model, minimum width=84pt, minimum height=76pt] (model4) at (376.2pt,0) {};
\node[below=6pt, align=center] at (model4.south) {layer $M$};
\node[swa, minimum width=0.7cm, minimum height=52pt] (attn4) at ($(model4.center) + (-21pt, 0)$) {\rotatebox{90}{Attention}};
\node[glu, minimum width=0.7cm, minimum height=52pt] (moe4) at ($(model4.center) + (21pt, 0)$) {\rotatebox{90}{MoE}};

\draw[-{Latex[length=1.4mm]}, line width=0.45pt] ($(model1.north) + (-7.1pt, 4.6pt)$) arc[start angle=230, end angle=-58, radius=11pt] node[pos=0.43, above=3pt] {loop $\times\, N$};
\draw[-{Latex[length=1.4mm]}, line width=0.45pt] ($(model2.north) + (-7.1pt, 4.6pt)$) arc[start angle=230, end angle=-58, radius=11pt] node[pos=0.43, above=3pt] {loop $\times\, N$};
\draw[-{Latex[length=1.4mm]}, line width=0.45pt] ($(model3.north) + (-7.1pt, 4.6pt)$) arc[start angle=230, end angle=-58, radius=11pt] node[pos=0.43, above=3pt] {loop $\times\, N$};
\draw[-{Latex[length=1.4mm]}, line width=0.45pt] ($(model4.north) + (-7.1pt, 4.6pt)$) arc[start angle=230, end angle=-58, radius=11pt] node[pos=0.43, above=3pt] {loop $\times\, N$};

\coordinate (left_input_start) at ($(model1.west) + (-31.85pt, 0)$);
\coordinate (right_output_end) at ($(model4.east) + (31.85pt, 0)$);
\node[anchor=east, align=center, font=\bfseries\large, overlay] at ($(left_input_start) + (-14pt, 0)$) {Layer-loop};
\draw[layerlink] (left_input_start) -- (attn1.west);
\draw[layerlink] (attn1.east) -- (moe1.west);
\draw[layerlink] (moe1.east) -- (attn2.west);
\draw[layerlink] (attn2.east) -- (moe2.west);
\draw[layerlink] (moe2.east) -- (attn3.west);
\draw[layerlink] (attn3.east) -- (moe3.west);
\draw[layerlink] (moe3.east) -- node[midway, fill=white, inner sep=1.5pt, font=\Large] {$\cdots$} (attn4.west);
\draw[layerlink] (attn4.east) -- (moe4.west);
\draw[layerlink] (moe4.east) -- (right_output_end);

\end{tikzpicture}
}

}
\par\vspace{5mm}
\noindent\makebox[\textwidth][c]{%
\raisebox{6mm}{%
\tikz\draw[gray!70, densely dashed, line width=0.6pt] (0,0) -- (\textwidth,0);
}%
}
\raisebox{5mm}{\scalebox{0.85}{
\centering
\resizebox{0.7\textwidth}{!}{

\begin{tikzpicture}

\node[model, minimum width=84pt, minimum height=76pt] (model1) at (0,0) {};
\node[below=6pt, align=center] at (model1.south) {layer 1};
\node[swa, minimum width=0.7cm, minimum height=52pt] (attn1) at ($(model1.center) + (-21pt, 0)$) {\rotatebox{90}{Attention}};
\node[glu, minimum width=0.7cm, minimum height=52pt] (moe1) at ($(model1.center) + (21pt, 0)$) {\rotatebox{90}{MoE}};

\node[model, minimum width=84pt, minimum height=76pt] (model2) at (123.1pt,0) {};
\node[below=6pt, align=center] at (model2.south) {layer 2};
\node[swa, minimum width=0.7cm, minimum height=52pt] (attn2) at ($(model2.center) + (-21pt, 0)$) {\rotatebox{90}{Attention}};
\node[glu, minimum width=0.7cm, minimum height=52pt] (moe2) at ($(model2.center) + (21pt, 0)$) {\rotatebox{90}{MoE}};

\node[model, minimum width=84pt, minimum height=76pt] (model3) at (246.2pt,0) {};
\node[below=6pt, align=center] at (model3.south) {layer 3};
\node[swa, minimum width=0.7cm, minimum height=52pt] (attn3) at ($(model3.center) + (-21pt, 0)$) {\rotatebox{90}{Attention}};
\node[glu, minimum width=0.7cm, minimum height=52pt] (moe3) at ($(model3.center) + (21pt, 0)$) {\rotatebox{90}{MoE}};

\node[model, minimum width=84pt, minimum height=76pt] (model4) at (376.2pt,0) {};
\node[below=6pt, align=center] at (model4.south) {layer $M$};
\node[swa, minimum width=0.7cm, minimum height=52pt] (attn4) at ($(model4.center) + (-21pt, 0)$) {\rotatebox{90}{Attention}};
\node[glu, minimum width=0.7cm, minimum height=52pt] (moe4) at ($(model4.center) + (21pt, 0)$) {\rotatebox{90}{MoE}};

\coordinate (left_input_start) at ($(model1.west) + (-31.85pt, 0)$);
\coordinate (right_output_end) at ($(model4.east) + (31.85pt, 0)$);
\coordinate (skip_input) at ($(left_input_start)!0.5!(attn1.west)$);
\coordinate (skip_output) at ($(moe4.east)!0.5!(right_output_end)$);
\node[anchor=east, align=center, font=\bfseries\large, overlay] at ($(left_input_start) + (-14pt, 0)$) {Model-loop};
\draw[layerlink] (left_input_start) -- (attn1.west);
\draw[layerlink] (attn1.east) -- (moe1.west);
\draw[layerlink] (moe1.east) -- (attn2.west);
\draw[layerlink] (attn2.east) -- (moe2.west);
\draw[layerlink] (moe2.east) -- (attn3.west);
\draw[layerlink] (attn3.east) -- (moe3.west);
\draw[layerlink] (moe3.east) -- node[midway, fill=white, inner sep=1.5pt, font=\Large] {$\cdots$} (attn4.west);
\draw[layerlink] (attn4.east) -- (moe4.west);
\draw[layerlink] (moe4.east) -- (right_output_end);
\coordinate (model_loop_bottom_left) at ($(skip_input) + (0,-72pt)$);
\coordinate (model_loop_bottom_right) at ($(skip_output) + (0,-72pt)$);
\draw[-{Latex[length=2.1mm]}, black, densely dashed, line width=0.32pt] (skip_output) -- (model_loop_bottom_right) -- (model_loop_bottom_left) -- (skip_input);
\node[below=3pt] at ($(model_loop_bottom_left)!0.5!(model_loop_bottom_right)$) {loop $\times\, N$};

\end{tikzpicture}
}
}}
\vspace{3mm}
\caption{Illustration of the contrast between \textbf{layer-loop} and \textbf{model-loop}. The top panel shows layer-loop, the loop schedule adopted by Loopie, in which each Attention/MoE layer is applied recurrently before passing its output to the next layer. We demonstrate that this schedule achieves better performance on MoE backbones while also training more efficiently. The bottom panel shows the traditional whole-model recurrence pattern, which we call model-loop, where the entire layer stack is traversed and then repeated. This pattern is used by prior looped models such as Ouro and Huginn \citep{zhu2025scalinglatent,geiping2025scalinglatent}.}
\label{fig:layer-vs-model-loop}
\end{figure*}
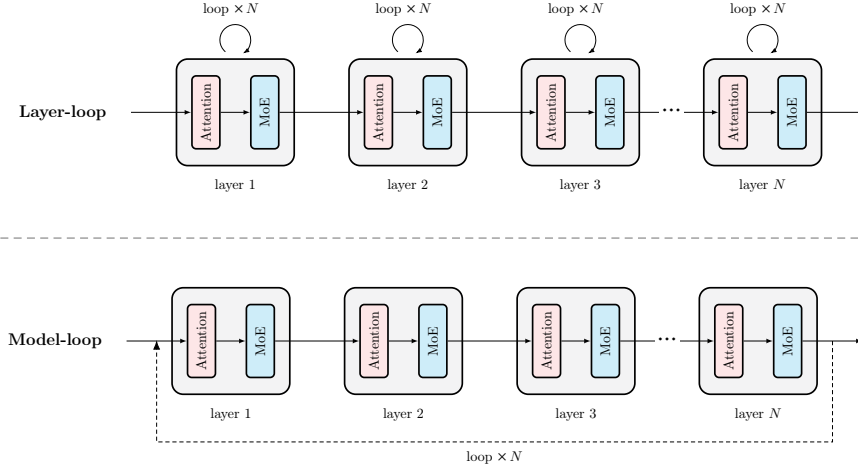

\section{The Loopie Series}

This section presents the Loopie Series, from its architecture to the evidence for its scalability. We begin by defining Loopie's layer-loop recurrence pattern and contrasting it with the classic model-loop pattern used in prior looped language models. We then motivate the design under a fixed pre-training compute budget, where models that use recurrent depth must be compared against vanilla Transformers that spend the same compute on ordinary non-recurrent capacity. Building on this motivation, we introduce the Loopie Recipe, a compute-matched scaling recipe for choosing stored width, stored depth, and recurrent depth. We next present the main compute-matched results, examine whether the advantage persists across model scales, and isolate the contribution of the layer-loop pattern through ablations. Finally, we discuss why Loopie uses only two recurrent steps.

\subsection{Architecture}
\label{subsec:architecture}

Loopie largely follows the Qwen3-MoE architecture. In particular, its backbone is a decoder-only Mixture-of-Experts Transformer, and its attention mechanism, sparsity pattern, and other architectural details remain the same as those of the Qwen3-MoE family. Detailed architectural specifications are provided in Appendix~\ref{arch_details}. The key architectural difference lies in how recurrent computation is applied. Instead of simply repeating the entire model multiple times, Loopie adopts a different recurrence pattern that we call \textbf{layer-loop}.

Prior looped language models, such as Ouro and Huginn \citep{zhu2025scalinglatent,geiping2025scalinglatent}, mainly use what we call \textbf{model-loop}. In model-loop recurrence, the entire Transformer stack is unrolled recurrently: for a model with three layers and two loop steps, the computation order is
\[
\text{Layer 1} \rightarrow \text{Layer 2} \rightarrow \text{Layer 3}
\rightarrow \text{Layer 1} \rightarrow \text{Layer 2} \rightarrow \text{Layer 3}.
\]
By contrast, Loopie uses \textbf{layer-loop}, in which each layer is applied recurrently before the computation moves to the next layer. For the same three-layer, two-step example, the computation order becomes
\[
\text{Layer 1} \rightarrow \text{Layer 1} \rightarrow
\text{Layer 2} \rightarrow \text{Layer 2} \rightarrow
\text{Layer 3} \rightarrow \text{Layer 3}.
\]
That is, each block performs local iteration on the hidden states and only then passes the resulting recurrent representation to the next layer. Figure~\ref{fig:layer-vs-model-loop} illustrates the difference.

\subsection{Why Layer-Loop?}
Layer-loop is not merely a different ordering of recurrent computation; it changes where iterative refinement happens inside the model and therefore affects scaling behavior, execution efficiency, and the nature of parameter sharing across effective depth. We adopt layer-loop because it offers three advantages that are especially important for large-scale pre-training.

\begin{figure}[!htbp]
    \centering
    \includegraphics[width=0.6\linewidth]{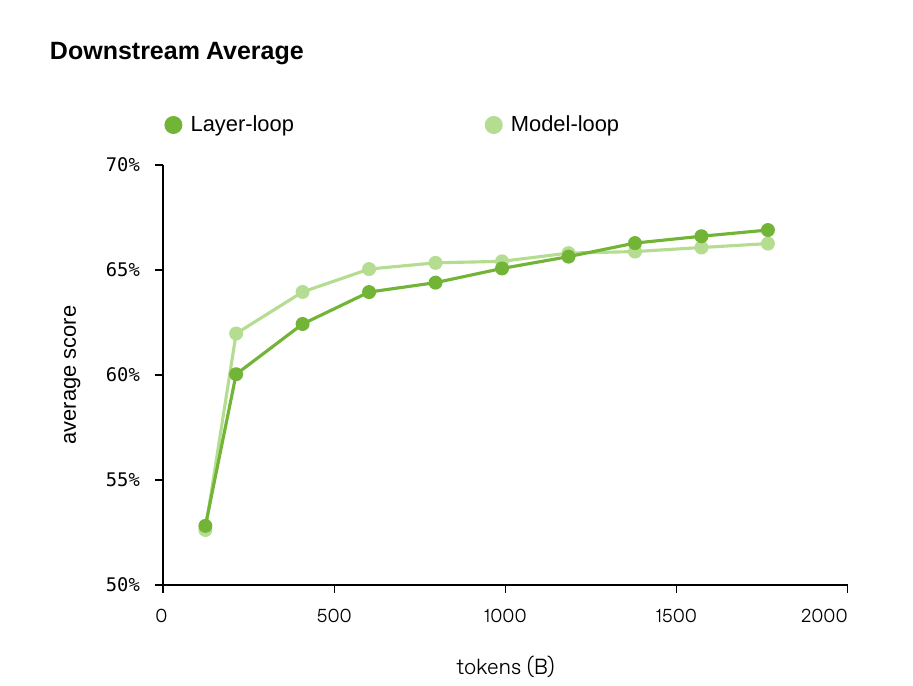}
    \caption{Comparison of the average downstream benchmark scores between the layer-loop and model-loop variants of Loopie-6B-A0.6B. Although layer-loop lags behind model-loop early in training, it surpasses model-loop later.}
    \label{fig:layerloop}
\end{figure}

\paragraph{Better empirical scaling.}
Layer-loop achieves better performance than model-loop in our pre-training
experiments. As shown in Figure~\ref{fig:layerloop}, layer-loop initially
trails model-loop slightly on downstream benchmarks. However, it overtakes
model-loop after approximately \(1.2\) trillion training tokens and improves
more rapidly thereafter. The early advantage of
model-loop therefore does not persist as the training budget increases.

\FloatBarrier

\paragraph{Infrastructure-friendly execution.}
Although layer-loop and model-loop have the same nominal number of layer applications and theoretical FLOPs, layer-loop provides better execution locality. Repeated applications of the same layer are adjacent in both the forward and backward computation graphs, which shortens the reuse distance for parameters and gradient contributions and simplifies activation checkpointing and gradient accumulation, especially under parameter sharding or offloading. This locality is particularly beneficial for pipeline parallelism because all recurrent applications of a layer remain within the same pipeline stage before activations are transferred onward. In contrast, model-loop requires each microbatch to traverse the entire pipeline repeatedly and routes the output of the final stage back to the first stage at each loop boundary, introducing cyclic dependencies that may complicate scheduling, increase communication overhead and the number of pipeline bubbles, and reduce device utilization.

\paragraph{Natural parameter-sharing pattern.}
For example, consider a 48-layer model such as Qwen3-30B-A3B with two model-loop steps. The third physical layer is applied at effective depths \(3\) and \(48 + 3 = 51\). These two invocations occur after markedly different amounts of preceding computation and therefore receive hidden states at widely separated effective depths. Prior analyses suggest that Transformer representations are organized nonuniformly across depth: lower and final layers can differ substantially from the comparatively homogeneous middle layers, while different linguistic abstractions tend to become most accessible at different stages of the network \citep{sun2025transformer,tenney2019bert,jawahar2019bert}. These observations do not directly establish gradient conflict, but they suggest that hidden states at widely separated effective depths need not place identical functional demands on a shared transformation. Thus, model-loop asks a single parameter set to accommodate potentially heterogeneous depth-dependent roles. In contrast, layer-loop reuses a layer at adjacent effective depths, resulting in a more local and potentially more coherent parameter-sharing pattern.

\subsection{Motivation}
\label{subsec:motivation}

Despite their conceptual appeal, looped Transformers have historically been studied in settings that do not fully reflect the constraints of modern large-scale language model pre-training. In particular, much of the prior work has focused on dense, weight-shared, or recurrent Transformer variants and has evaluated recurrence primarily as a mechanism for improving parameter efficiency~\citep{zhu2025scalinglatent,geiping2025scalinglatent}. This leaves two important issues underexplored.

First, modern frontier language models increasingly rely on Mixture-of-Experts architectures, as exemplified by the recent Qwen3, Kimi K2.5, GLM-5, DeepSeek-V4, and MiniMax-M2 model families~\citep{qwen3,kimi2026kimi25,glm5team2026glm5,deepseekai2026deepseekv4,minimax2026m2}. MoE models expand total model capacity while keeping the number of active parameters per token relatively small, making them substantially more favorable than dense models under both training and inference compute constraints~\citep{shazeer2017outrageously,fedus2022switch,du2022glam,abnar2025parameters}. A practical looped architecture should therefore be compatible with MoE scaling rather than be restricted to dense backbones, because such a restriction would limit its ability to scale to larger model sizes.

Second, pre-training compute is often the dominant constraint in large-model training. It determines how large a model can be, how many tokens it can see, and how many experimental variants can be trained~\citep{kaplan2020scaling,hoffmann2022training}. This creates a key difficulty for looped Transformers: recurrent computation is not free. If a model is looped $N$ times during pre-training, its training compute is also multiplied by approximately $N$. Therefore, a looped model should not merely outperform a vanilla Transformer with the same stored parameter count. To justify recurrence as a scaling strategy, it must compete with vanilla Transformers trained under the same pre-training compute budget.

This requirement is challenging. Prior work suggests that, under fixed-FLOP comparisons, a looped model can have less non-recurrent capacity than a standard Transformer baseline~\citep{csordas2024moeut,frey2026dualpath}. This is consistent with the conventional view that recurrence mainly improves parameter efficiency but not necessarily compute efficiency~\citep{lan2019albert,takase2023lessons}. As a result, prior looped Transformers have not yet provided a clear path toward flagship models at the trillion-parameter scale with frontier-level performance.

Loopie is designed to address this gap. Our central observation is that recurrence becomes competitive under a fixed compute budget only when it is paired with an appropriate recurrence-width-depth trade-off. We formalize this principle as the \textbf{Loopie Recipe}: rather than treating loops as a direct substitute for additional parameters, one should jointly choose stored width, stored depth, and recurrent depth so that the resulting model maximizes performance under a fixed pre-training compute budget. This changes the role of recurrence from a parameter-saving device into a compute-matched scaling mechanism.

\subsection{Loopie Recipe}
\label{subsec:loopie-recipe}

We now describe the empirical, hardware-aware scaling procedure used to
instantiate Loopie-20B-A2B. Starting from a strong non-recurrent MoE
reference, the \textbf{Loopie Recipe} comprises three steps:
\begin{tcolorbox}[colback=blue!4,colframe=blue!35!black,boxrule=0.5pt,arc=4pt,left=6pt,right=6pt,top=5pt,bottom=5pt]
(i) constructing a recurrent seed model by halving the number of stored layers; \\
(ii) executing each stored layer twice using \textit{layer-loop}; and \\
(iii) using the resulting memory headroom to double the per-device
microbatch size, then reinvesting the measured training efficiency gain
into additional model capacity while keeping the optimizer-step time
approximately matched to the reference model.
\end{tcolorbox}

In this work, we match models by realized pre-training cost rather than by exact theoretical FLOPs. For every comparison, we fix the hardware allocation, sequence length, number of tokens per optimizer step, activation-checkpointing policy, optimizer, and training data. Architectures may use different per-device microbatch sizes according to their measured memory footprints. When the microbatch size is increased, the number of gradient-accumulation steps is reduced proportionally so that the number of tokens per optimizer step remains unchanged.

We select the Loopie configuration whose measured end-to-end optimizer-step time most closely matches the non-recurrent reference. Since both models use the same token budget and number of optimizer updates, this also approximately matches total wall-clock training cost.

The models are not matched by their theoretical FLOP counts. Loopie performs more nominal computation per token, but the additional work is offset by the higher realized efficiency. We refer to this operationally as a compute-matched comparison. Analytical FLOP and memory models are used only to construct and
interpret candidate configurations. The final architecture is selected
using measured end-to-end training time in Megatron-LM~\citep{megatron-lm}.

The measured efficiency gain comes solely from the reduced stored depth, which lowers activation memory and enables us to double the per-device microbatch size while halving the number of gradient-accumulation steps.

\paragraph{Activation memory and microbatch efficiency.}
Under the checkpointing implementation used in our experiments, all
recurrent applications of a stored layer are enclosed in the same
checkpointed unit. Consequently, the dominant activation memory term
scales with stored depth rather than executed depth. The activation memory $M_{\mathrm{act}}$ during training scales as~\citep{MLSYS2023_80083951,3433701.3433727}
\[
M_{\mathrm{act}}
\propto
s\,bDL,
\]
where $D$ is the hidden dimension, $L$ is the number of stored layers,
$b$ is the per-device microbatch size, and $s$ is the sequence
length. The recurrent applications increase the amount of executed computation through the loop count $R$ but do not introduce $R$ independently stored sets of layer-boundary activations under this checkpointing scheme.

For a fixed global batch size \( B \), with per-device microbatch size \( b \) and \( g \) gradient-accumulation steps, we have
\[
g = \frac{B}{b}.
\]
Reducing $DL$ creates memory headroom that can be used to increase
$b$. Because the number of gradient-accumulation steps $g$ is
reduced proportionally, the global batch size remains fixed:
\[
b_{1}
=
2b_{0},
\qquad
g_1
=
\frac{g_0}{2}.
\]
The larger microbatch exposes more parallel work to each kernel and
reduces the number of gradient-accumulation micro-steps required for
each optimizer update.

For a fixed architecture, the corresponding microbatch efficiency gain
is measured directly as
\[
S_{\mathrm{mb}}
=
\frac{
t_{\mathrm{step}}(D,L,R;b,g)
}{
t_{\mathrm{step}}(D,L,R;2b,g/2)
}.
\]
This measured quantity is the efficiency factor used by the
Loopie Recipe.

We use a Qwen3-like 30B-A3B MoE
Transformer~\citep{qwen3} as the non-recurrent reference, with
\[
D_0 = 2048,
\qquad
L_0 = 48,
\qquad
R_0 = 1.
\]
The first step is to construct a recurrent seed model by halving the number of
stored layers and setting $R=2$:
\[
D = 2048,
\qquad
L = 24,
\qquad
R = 2.
\]
The leading-order pre-training compute~\citep{hoffmann2022training} scales as
\[
C(D,L,R) \propto L D^2 R.
\]
At a fixed width, this transformation preserves the number of Transformer block executions:
\[
LR
=
24\cdot 2
=
48
=
L_0R_0,
\]
and therefore preserves the leading-order Transformer block compute proxy:
\[
LRD^2
=
24\cdot 2\cdot 2048^2
=
48\cdot 2048^2.
\]

At the same per-device microbatch size, the dominant activation memory term is reduced by half:
\[
\frac{DL}{D_0L_0}
=
\frac{2048\cdot 24}{2048\cdot 48}
=
0.5.
\]
The recurrent seed model therefore performs approximately the same
leading-order Transformer block work as the reference while requiring substantially less memory for stored activations~\citep{megatron-lm,3433701.3433727}.

The second step is to spend the resulting microbatch efficiency on
additional model capacity. To satisfy architectural and hardware alignment constraints, we restrict the candidate hidden sizes to multiples of 128. We sweep feasible $(D,L)$ configurations around the recurrent seed model and retain candidates that support a microbatch size twice that of the reference.

Table~\ref{tab:loopie-recipe} lists the candidate configurations. We report the normalized leading-order compute proxy:
\[
\widehat{C}
=
\frac{LRD^2}{48\cdot 2048^2},
\]
and the activation memory proxy at the reference microbatch size,
\[
\widehat{M}_{\mathrm{act}}
=
\frac{DL}{48\cdot 2048}.
\]
These quantities are used to describe the candidates, not to predict
their final optimizer-step times.

\begin{table}[t]
    \centering
    \small
    \caption{
    Candidate configurations generated from the Qwen3-like 30B-A3B
    reference. $\widehat{C}$ is the normalized leading-order Transformer block work
    proxy, and $\widehat{M}_{\mathrm{act}}$ is the normalized
    activation memory proxy at a fixed per-device microbatch size.
    Final compute matching is based on measured optimizer-step wall-clock
    time.
    }
    \label{tab:loopie-recipe}
    \begin{tabular}{lrrrrr}
        \toprule
        Configuration
        & $D$
        & $L$
        & $R$
        & $\widehat{C}$
        & $\widehat{M}_{\mathrm{act}}$ \\
        \midrule
        Qwen3 30B-A3B
        & 2048 & 48 & 1 & $1.00\times$ & $1.00\times$ \\
        \midrule
        Seed Loopie
        & 2048 & 24 & 2 & $1.00\times$ & $0.50\times$ \\
        Loopie candidate 1
        & 2176 & 25 & 2 & $1.18\times$ & $0.55\times$ \\
        Loopie candidate 2
        & 2304 & 27 & 2 & $1.42\times$ & $0.63\times$ \\
        Loopie candidate 3
        & 2432 & 28 & 2 & $1.65\times$ & $0.69\times$ \\
        \bottomrule
    \end{tabular}
\end{table}

We conduct large-scale benchmarking of each candidate configuration in Megatron-LM, following the matching protocol described above. For the reference model and each candidate model, we jointly search over tensor parallelism, expert parallelism, and microbatch size. We then select the candidate whose measured optimizer-step time is closest to that of the non-recurrent reference model:
\[
D_1 = 2304,
\qquad
L_1 = 27,
\qquad
R_1 = 2,
\]
which defines Loopie-20B-A2B.

At the reference per-device microbatch size, the dominant
activation memory proxy of Loopie-20B-A2B relative to the reference is
\[
\frac{D_1L_1}{D_0L_0}
=
\frac{2304\cdot 27}{2048\cdot 48}
\approx
0.633.
\]
This expression is a scaling proxy rather than an exact peak-memory
equation. Candidate feasibility is therefore determined using the
measured full memory footprint, which also includes parameters,
optimizer states, temporary buffers, and communication workspaces. The
measured memory profile of Loopie-20B-A2B permits the per-device
microbatch size to be doubled:
\[
b_{1}
=
2b_{0},
\qquad
g_1
=
\frac{g_0}{2}.
\]
This leaves the number of tokens per optimizer step unchanged. The selected model has a normalized leading-order Transformer block compute proxy of
\[
\frac{\widehat{C}_1}{\widehat{C}_0}
=
\frac{
27\cdot 2\cdot 2304^2
}{
48\cdot 2048^2
}
\approx
1.424.
\]
Loopie-20B-A2B matches the reference optimizer-step time only after
changing the schedule from $(b_0, g_0)$ to $(2b_0, g_0/2)$; thus, the measured efficiency gain is entirely attributable to the doubled
microbatch size.

The compute-matching criterion is therefore the directly measured relation:
\[
t_{\mathrm{step}}
\left(
D_1,L_1,R_1;
2b_{0},g_0/2
\right)
\approx
t_{\mathrm{step}}
\left(
D_0,L_0,R_0;
b_{0},g_0
\right),
\]
rather than an analytical prediction based on $\widehat{C}$. The
leading-order compute proxy omits lower-order operators, routing and
communication costs, optimizer overhead, kernel-launch overhead, and
the effect of the microbatch schedule on realized hardware
utilization. It is therefore used to characterize nominal work, not to replace end-to-end timing. We evaluated the Loopie Recipe across multiple GPU platforms and observed consistent infrastructure-level gains.

The Loopie Recipe therefore does not claim equality in analytical FLOP estimates. It uses layer-loop recurrence to reduce the stored memory footprint, converts the resulting memory headroom into a doubled per-device microbatch size, spends the measured microbatch efficiency gain on additional model capacity, and selects the final architecture using measured end-to-end optimizer-step time.

\begin{figure}[!htbp]
    \centering
    \includegraphics[width=\linewidth]{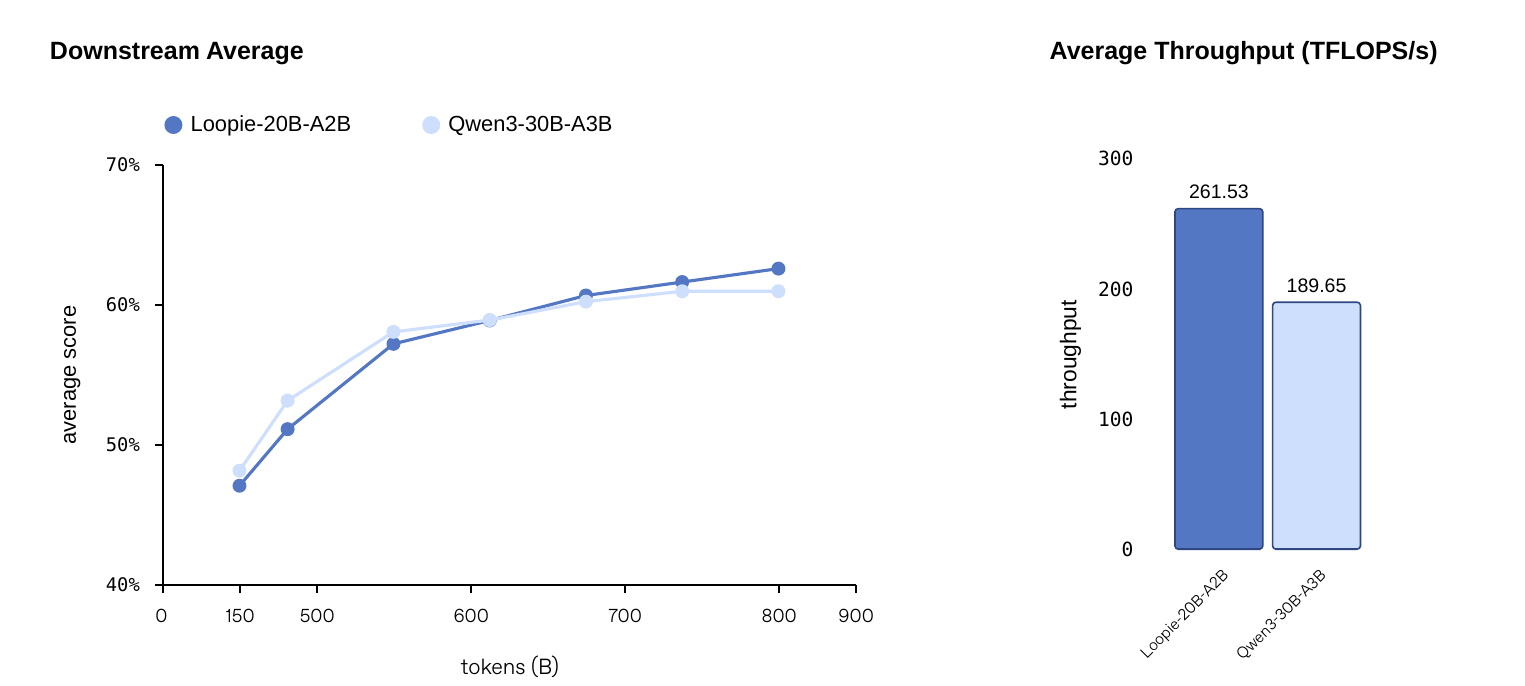}
    \caption{Comparison of Loopie-20B-A2B with our reproduction of Qwen3-30B-A3B under matched per-step pre-training wall-clock time in Megatron-LM. The left panel reports the average score across eight downstream benchmarks, while the right panel reports the highest average pre-training throughput achieved for each model during large-scale benchmarking. In the right figure, Loopie-20B-A2B uses EP = 8 and MBS = 2, while Qwen3-30B-A3B uses EP = 8 and MBS = 1. These configurations achieve the highest throughput in a large-scale TP/PP/EP/MBS grid search; the optimal settings may vary across different GPU.}
    \label{fig:ablation-qwen3}
\end{figure}
\FloatBarrier

\subsection{Results}

To test this principle, we compare Loopie against a strong non-recurrent baseline based on the Qwen3-MoE design. Specifically, we train a vanilla 30B-A3B MoE Transformer with a Qwen3-like architecture on 800 billion tokens under the same pre-training compute budget used for Loopie-20B-A2B. As shown in Figure~\ref{fig:ablation-qwen3}, Loopie initially lags behind the larger vanilla baseline during the early phase of training. However, after roughly 600 billion tokens of pre-training, Loopie-20B-A2B overtakes the compute-matched baseline and maintains a consistent advantage thereafter. This result suggests that, with the right scaling recipe, looped MoE models can achieve stronger final performance by using recurrent computation than by allocating the same compute to a larger vanilla Transformer.

\subsection{Scalability}
\label{subsec:scalability}

A practical recurrent architecture should not only outperform a single compute-matched baseline but also remain effective as the model is scaled. Many architectural ideas show promising results at one small scale yet fail at larger scales for various reasons. We therefore evaluate whether the Loopie design preserves its advantage across a sequence of increasingly large pre-training runs.

\begin{figure}[!htbp]
    \centering
    \includegraphics[scale=0.6]{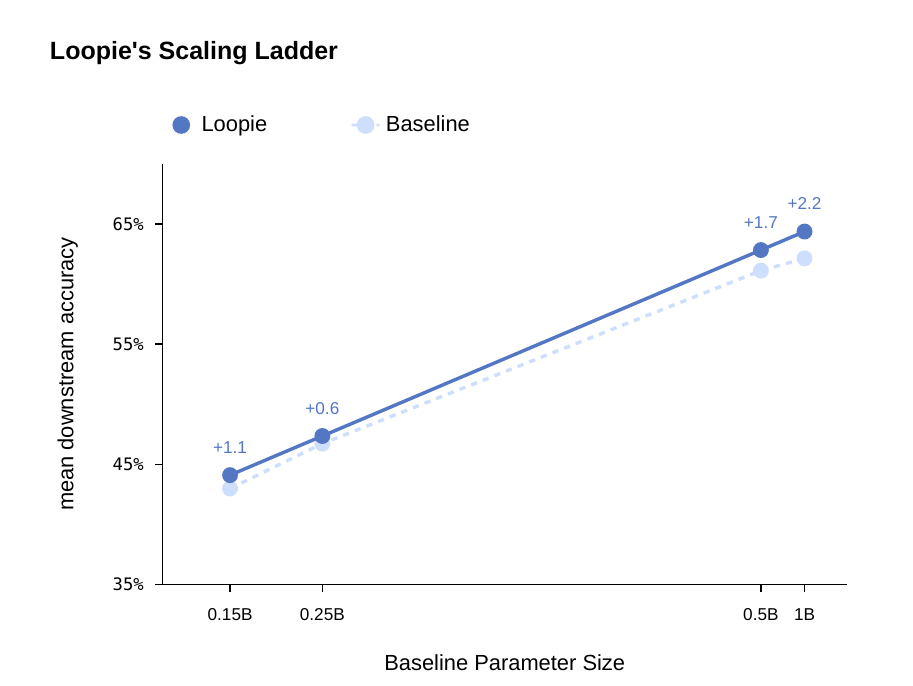}
    \caption{Scaling ladder for Loopie. The horizontal axis denotes the parameter count of the non-recurrent baseline, and the vertical axis reports the average score across eight downstream benchmarks. The Loopie models are not plotted at the parameter counts indicated on the horizontal axis; instead, each Loopie model is sized according to the Loopie Recipe so that its per-step pre-training wall-clock time exactly matches that of the corresponding baseline.}
    \label{fig:loopie-scalability}
\end{figure}

We construct a scaling ladder consisting of four non-recurrent MoE baselines and four compute-matched Loopie models. Each Loopie model is obtained by transforming a non-recurrent MoE baseline using the Loopie Recipe. At each rung, the Loopie model uses two layer-loop steps, and its stored width and depth are chosen according to the same compute-matched principle described in Section~\ref{subsec:loopie-recipe}. The goal of this ladder is not to match the stored parameter count but to match the effective pre-training compute of the corresponding vanilla baseline.
Thus, the Loopie models contain fewer stored parameters while using recurrence to increase effective depth.

For each reference model, we train on a token count equal to 1000× its active MoE parameter count, ensuring that each model is sufficiently overtrained and that its downstream metrics have largely stabilized. We use the same token budget for the corresponding Loopie models, even though these models have fewer active parameters.

For the smallest rung, we train the 0.15B vanilla baseline and its compute-matched 0.10B Loopie counterpart on 150B tokens, placing the models in a heavily overtrained regime relative to standard Chinchilla-style compute-optimal prescriptions. For the 0.25B and 0.50B rungs, we train on 250B and 500B tokens, respectively. For the 1B vanilla baseline and the corresponding 0.70B Loopie model, we also train on 500B tokens due to limited compute. The architectural details for this scaling ladder are provided in Appendix Table~\ref{tab:loopie-scaling-ladder}.

Figure~\ref{fig:loopie-scalability} summarizes the resulting scaling behavior.
Across all four rungs, Loopie consistently outperforms its compute-matched vanilla counterpart.
More importantly, the gap does not vanish as model size increases.
This suggests that Loopie's advantage persists across a meaningful scaling ladder rather than being merely a small-model artifact.
The result supports the central design hypothesis of the Loopie Series: recurrent layer-loop computation can be converted into scalable modeling gains when stored width, stored depth, and recurrent depth are jointly chosen under a fixed compute budget.

Taken together, the scaling ladder provides evidence that Loopie remains effective beyond an isolated compute-matched comparison.
As the baseline parameter count grows from 0.15B to 1B, Loopie models continue to deliver stronger downstream accuracy under matched pre-training budgets.
This behavior is important for large-scale deployment: it indicates that the layer-loop design and the Loopie Recipe are compatible with progressive scaling rather than being limited to a narrow model-size regime.

\subsection{Ablations}
\label{subsec:ablations}

\begin{figure}[H]
    \centering
    \includegraphics[width=0.6\linewidth]{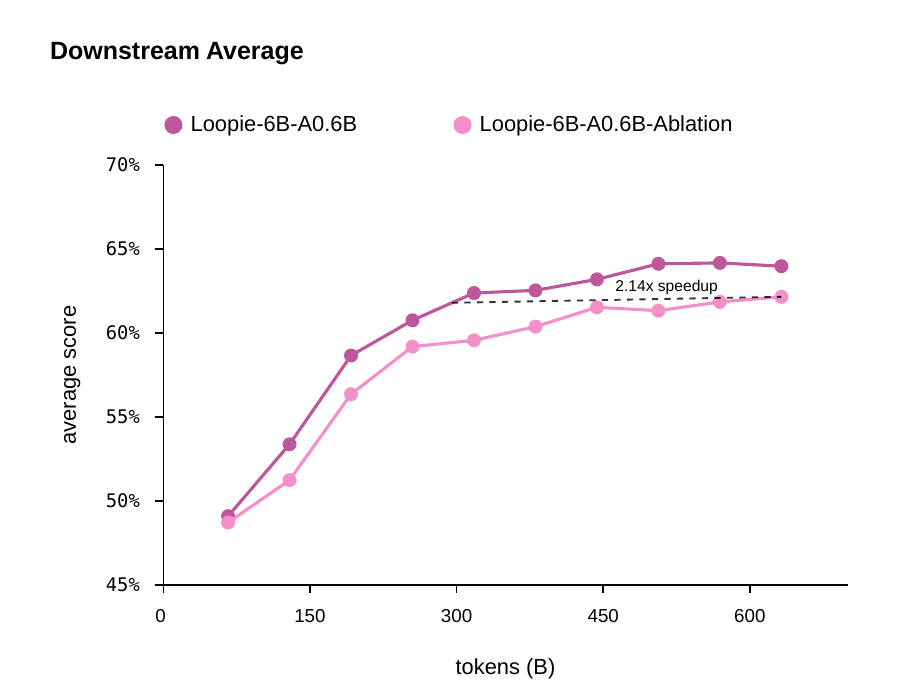}
    \caption{Layer-loop ablation for Loopie-6B-A0.6B. We report the average score across eight downstream benchmarks for Loopie-6B-A0.6B and Loopie-6B-A0.6B-Ablation. Loopie-6B-A0.6B-Ablation matches Loopie-6B-A0.6B in all architectural details except for the removal of layer-loop recurrence.}
    \label{fig:ablation-layerloop-6b}
\end{figure}

In addition to comparing Loopie against a compute-matched vanilla MoE baseline, we further isolate the contribution of the \emph{layer-loop} recurrence schedule. To do so, we conduct a controlled experiment by training a 6B-A0.6B MoE model with the same backbone, optimizer, data mixture, and token budget as Loopie-6B-A0.6B but with the layer-loop pattern removed. This ablated model therefore preserves the overall looped computation budget while testing whether the ordering of recurrent computation is itself important.

Figure~\ref{fig:ablation-layerloop-6b} shows the resulting average score across eight downstream benchmarks. The layer-loop pattern substantially improves downstream performance relative to the matched ablation. Importantly, because neither the number of active parameters nor the overall computation budget increases, the improvement cannot simply be attributed to additional computation.

This result supports the architectural choice made in Loopie. While recurrence increases effective depth, the way recurrence is scheduled matters: naively looping computation is insufficient to achieve high compute efficiency. These ablations suggest that Loopie's gains arise not only from using recurrence but from using recurrence in a form that is aligned with the hierarchical structure of Transformer representations.

\subsection{Why Only Two Loop Steps?}
\label{subsec:why-two-loops}

Many recurrent language model designs use substantially deeper unrolling. For example, prior looped or latent-recurrent models often use 4, 16, or even more than 30 recurrent steps \citep{zhu2025scalinglatent,geiping2025scalinglatent}. These settings are useful for studying recurrence as a general mechanism for iterative computation. Loopie targets a different regime: large-scale language model pre-training under a fixed compute budget. In this regime, the number of loop steps is not merely an architectural hyperparameter; it is a direct allocation of pre-training FLOPs.

At a fixed stored parameter count and a fixed number of optimizer steps, increasing the number of loop steps usually improves the training curve. However, this comparison is not compute-matched. For a model with $R$ loop steps, the per-token training cost scales approximately linearly with $R$, up to MoE routing and implementation constants:
\[
\mathcal{C}(D,L,R)
\propto
L D^2 R .
\]
Under the same pre-training budget, the model must either be trained on fewer tokens, use a smaller stored architecture, or be compared against a stronger non-recurrent model that spends the same compute on ordinary Transformer capacity. The relevant question is thus not whether a model with $R=4$ outperforms one with $R=2$ at the same stored size but whether the marginal gain from additional recurrence exceeds the gain that could be obtained by reallocating the same FLOPs to width, depth, or data.

\begin{figure}[t]
    \centering
    \includegraphics[width=\linewidth]{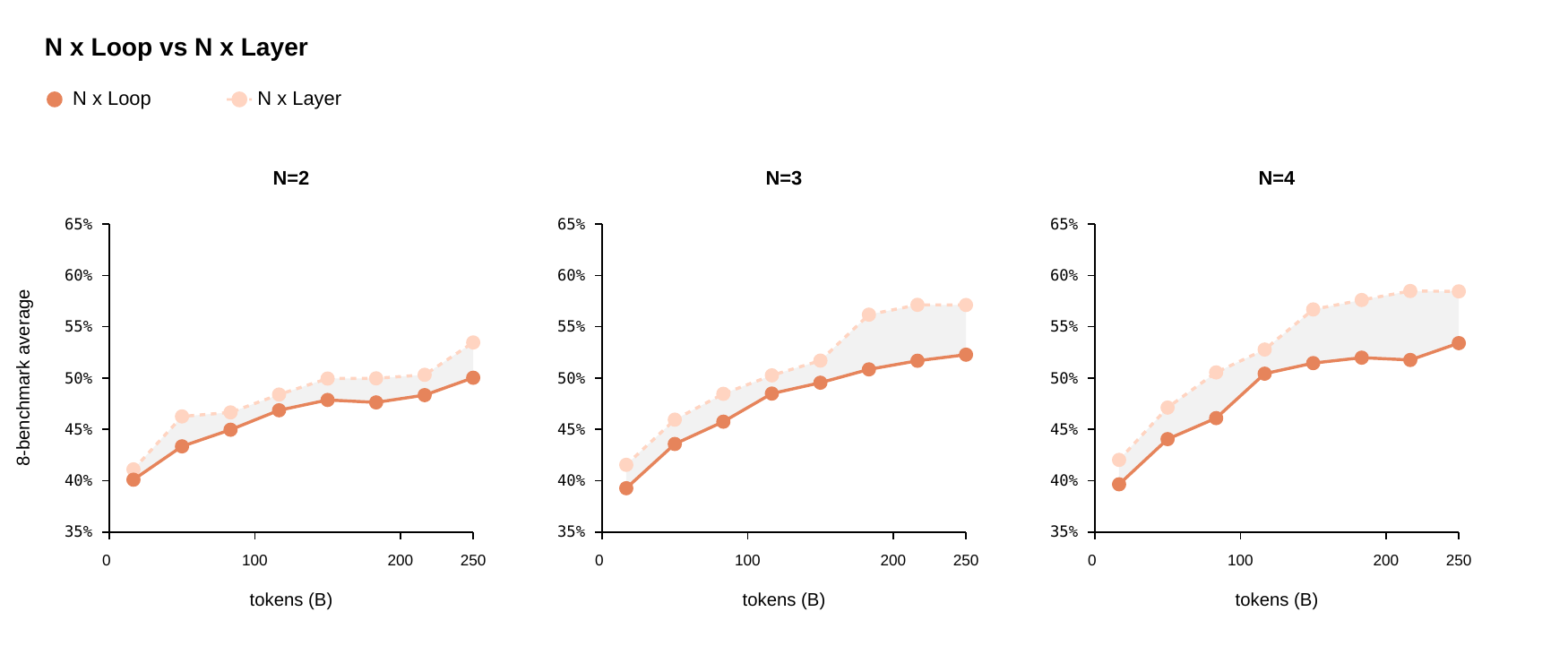}
    \caption{Comparison between $N\times$ layer-loop and $N\times$ stored-layer scaling. We report the average score across eight downstream benchmarks after training each model on 250B tokens while keeping all other architectural details fixed. The shaded gray region denotes the gap between the $N\times$ layer baseline and the $N\times$ loop model.}
    \label{fig:loop-count-sweep}
\end{figure}

Our loop-count sweep suggests that this marginal return decays rapidly in the large-scale pre-training regime. To isolate whether the gain comes from recurrence rather than simply adding more non-shared layers, we compare a model that uses $N\times$ layer-loop steps with one that has $N\times$ as many stored layers, while holding all other architectural details fixed. Figure~\ref{fig:loop-count-sweep} reports the average score across eight downstream benchmarks after the models are trained on 250B tokens. The shaded gray region highlights the performance gap between the $N\times$ layer baseline and the $N\times$ loop model. For the $2\times$ comparison, we use a $2\times$ stored-layer model only as a conservative proxy for the previous compute-matched setting. Because this model is very small, with only 0.25B active parameters, precise compute matching is difficult under multiple practical constraints, so the actual training compute of the $2\times$ layer baseline is substantially higher than that of the $2\times$ loop model. Thus, the figure should not be read as evidence that $2\times$ looping is dominated by $2\times$ layer scaling; rather, it shows that, when comparing $N\times$ looping against $N\times$ stored-layer scaling, the marginal benefit of recurrence is largest at $R=2$.

We therefore choose $R=2$ for the Loopie Series. This is the smallest nontrivial recurrent setting: each layer performs one ordinary transformation followed by one local refinement step before passing its representation to the next layer. This permits recurrence to change the computation qualitatively while keeping the compute multiplier small enough for scalable pre-training. It also preserves training throughput and makes the comparison against Qwen3-like vanilla MoE baselines stringent: Loopie must outperform these baselines not by using a large amount of extra recurrent computation but by using a small amount of recurrence more effectively.

This choice should not be interpreted as claiming that larger loop counts are ineffective in isolation. Larger $R$ may be useful in settings where inference-time computation is cheap, where adaptive computation is available, or where the goal is to study recurrent reasoning rather than pre-training efficiency. For frontier-scale pre-training, however, the dominant constraint is total compute. Under this constraint, $R=2$ provides the best trade-off we observe between iterative refinement, throughput, and compute-matched scaling.

\section{Pre-Training}
\label{sec:pre-training}

We pre-train Loopie models in two stages. In the first stage, we perform large-scale pre-training from scratch on 3T tokens. In the second stage, we conduct high-quality data annealing using 1.26T tokens, with an emphasis on high-quality synthetic, STEM, code, mathematical reasoning, and web data. All Loopie models use the tokenizer from the Qwen3 model family~\citep{qwen3}. We conduct all pre-training using the Megatron-LM~\citep{megatron-lm} framework.

\subsection{Evaluation}
\label{sec:pre-training_eval}

We evaluate the pre-trained checkpoints using the LM Evaluation Harness framework~\citep{eval-harness}. For all experiments in Section~2, we report the mean score across the following eight benchmarks: ARC-Challenge~\citep{clark2018think}, ARC-Easy~\citep{clark2018think}, BoolQ~\citep{clark2019boolq}, CommonsenseQA~\citep{talmor2019commonsenseqa}, HellaSwag~\citep{zellers2019hellaswag}, MMLU~\citep{hendrycks2021measuring}, OpenBookQA~\citep{mihaylov2018openbookqa}, and WinoGrande~\citep{sakaguchi2020winogrande}.

\subsection{Initialization}
Let \(d\) denote the model width. We initialize the token embedding
matrix \(E\) and the language-modeling head \(W_{\mathrm{lm}}\)---or
the shared embedding/unembedding matrix when weight tying is used---as
\(E_{ij},(W_{\mathrm{lm}})_{ij}\sim\mathcal{N}(0,d^{-1})\). On the
input side, the looked-up embeddings are multiplied by \(\sqrt d\)
before entering the residual stream, so that the embedding shortcut has
\(O(1)\) activation scale, while the same \(d^{-1}\) variance ensures that
the initial logits \(h^\top w\) have \(O(1)\) variance for normalized hidden
states \citep{shazeer2018adafactor,takase2025spike}.
For hidden-to-hidden parameters inside Transformer blocks, including
attention projections and feed-forward layers, we use SmallInit:
\[
    W_{ij}\sim \mathcal{N}\!\left(0,\frac{2}{5d}\right),
    \qquad \operatorname{std}(W)=\frac{1}{\sqrt{2.5d}}.
\]
This scale corresponds to Xavier fan-in/fan-out initialization for a
Transformer FFN with an expansion ratio of \(4\), since \(d_{\rm in}+d_{\rm out}
=d+4d=5d\). We also apply this scale to attention projections to reduce the
scale of residual-branch sub-layers
\citep{glorot2010understanding,nguyen2019transformers}. Thus, we use
\(d^{-1/2}\) for embeddings/unembeddings, which determine the shortcut
and logit scales, but \((2.5d)^{-1/2}\) for block sub-layers, whose
Jacobians should remain small for stable pre-training
\citep{takase2025spike}.

\subsection{Learning Schedule}
We optimize all models with AdamW \citep{kingma2015adam,loshchilov2019decoupled}. We set the peak learning rate to \(5\times 10^{-4}\) for \textsc{Loopie-6B-A0.6B} and \(3\times 10^{-4}\) for \textsc{Loopie-20B-A2B}. We set the AdamW hyperparameters to \(\beta_1=0.9\), \(\beta_2=0.95\), and \(\epsilon=10^{-15}\), use a weight decay of \(0.1\), and apply global gradient-norm clipping at \(1.0\). We use a warmup-stable-only learning-rate schedule. After a specified number of warmup steps---6,000 steps for Loopie-20B-A2B and 2,000 steps for Loopie-6B-A0.6B---the learning rate reaches its peak value and then remains constant without decay. This decay-free stable phase keeps the effective update scale high after warmup, avoiding the diminished influence of late-stage high-quality data that can occur under annealed schedules \citep{wen2024understanding,luo2025learning,yano2026pretraining}. We also use a constant learning rate during the high-quality annealing phase in Stage 2 to maximize learning from the highest-quality annealing data~\citep{luo2025learning}. The global batch size is 1024, and the sequence length is 8192.

\subsection{Stage 1: Multi-Epoch High-Quality Pre-Training}
In Stage 1, we train both Loopie-20B-A2B and Loopie-6B-A0.6B on Nemotron-CC-v2-HQ~\citep{nemotron-cc}, a high-quality subset of Nemotron-CC. The corpus contains approximately 570B unique tokens, and we train for four epochs, totaling approximately 2.28T training tokens. This design is motivated by our prior observation that repeated training on high-quality data can be more effective than single-pass training on a larger corpus of lower average quality~\citep{gao2025makesdiffusionlanguagemodels}.

\subsection{Stage 2: High-Quality Annealing}
\begin{figure}[!htbp]
\centering
\includegraphics[width=\linewidth]{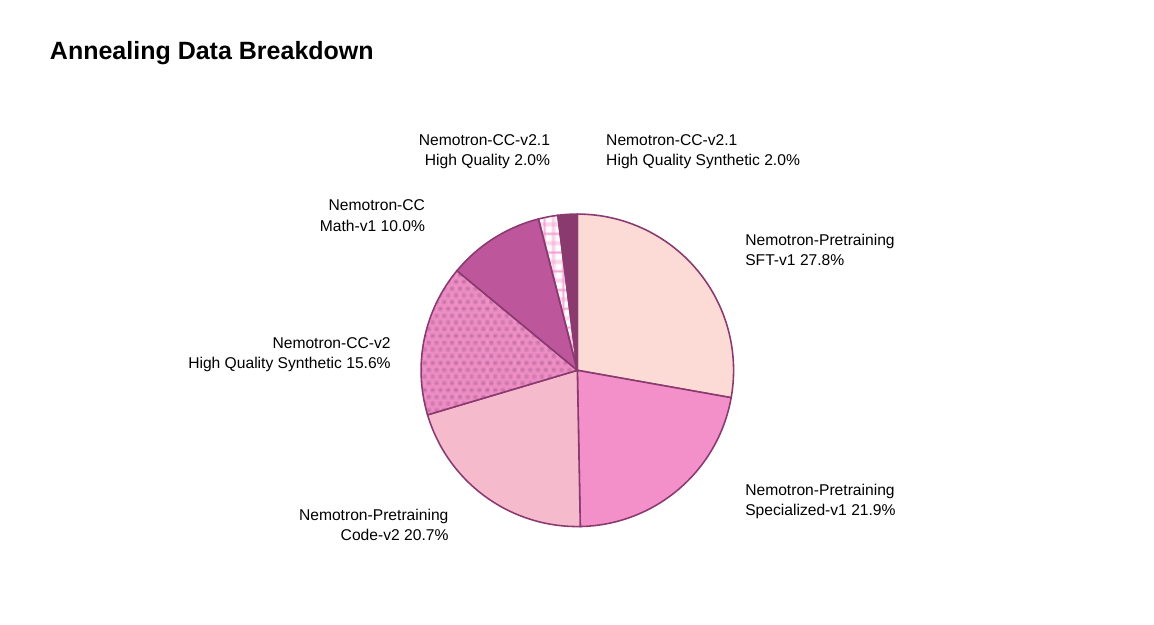}
\caption{Composition of the Stage-2 high-quality annealing data pool: multiple data sources make up the 1.26T-token annealing recipe.}
\label{fig:stage2_data_mix}
\end{figure}

In Stage 2, we construct a high-quality annealing mixture from Nemotron pre-training datasets. The resulting pool contains approximately 1.26T tokens. The mixture combines high-quality SFT-style data, synthetic reasoning data, code data, synthetic web data, and math data. In total, the pool contains approximately 1263B tokens: 351B from Nemotron-pre-training-SFT-v1 (27.8\%), 277B from Nemotron-pre-training-Specialized-v1 (21.9\%), 262B from a 60\% sample of Nemotron-pre-training-Code-v2 (20.7\%), 197B from a 16\% sample of Nemotron-CC-v2-HQ-Synthetic (15.6\%), 126B from Nemotron-CC-Math-v1 with quality scores $\geq 4$ (10.0\%), 25B from Nemotron-CC-v2.1-HQ (2.0\%), and 25B from Nemotron-CC-v2.1-HQ-Synthetic (2.0\%). Figure~\ref{fig:stage2_data_mix} visualizes these relative proportions.

\FloatBarrier

\paragraph{Nemotron-pre-training-SFT-v1.}
We include approximately 351B tokens from Nemotron-pre-training-SFT-v1, a diverse SFT-style dataset of synthetic and curated examples spanning STEM, academic, code, mathematics, and reasoning domains, with multilingual coverage. Its STEM component is expanded from high-quality math and science seeds through iterative generation with Qwen3 and DeepSeek models, producing harder and more varied questions with solutions. The dataset also contains academic question-answer pairs synthesized from undergraduate- and graduate-level texts, as well as MMLU-style general QA and fundamental reasoning data.

\paragraph{Nemotron-pre-training-Specialized-v1.}
We include approximately 277B tokens from Nemotron-pre-training-Specialized-v1, which comprises synthetic data for STEM reasoning, scientific coding, and cross-domain coding, as well as synthetic Wikipedia data and synthetic mathematics textbook data. The STEM reasoning component includes reasoning question-answer demonstrations generated from advanced scientific seed documents, while the scientific coding and cross-domain coding subsets introduce graduate- or research-level programming tasks with structured solutions. This data is intended to strengthen scientific reasoning, mathematical abstraction, and code generation during annealing.

\paragraph{Nemotron-pre-training-Code-v2.}
We randomly sample 60\% of Nemotron-pre-training-Code-v2, yielding approximately 262B tokens. This dataset combines recent GitHub source code, synthetic code-grounded question-answer data, student-teacher dialogues, code-review dialogues, and LLM-rewritten or transpiled source code. The rewriting and transpilation components are designed to improve downstream code generation by increasing stylistic diversity and exposing the model to semantically equivalent code variants.

\paragraph{High-quality synthetic web data.}
We randomly sample 16\% of Nemotron-CC-v2-High-Quality-Synthetic, yielding approximately 197B tokens. This dataset is derived from Nemotron-CC-v2 and contains English web-crawl documents augmented through synthetic rephrasing with Qwen3-30B-A3B. We further include 25B tokens from Nemotron-CC-v2.1-High-Quality-Synthetic, which extends the synthetic high-quality web corpus with newer Common Crawl snapshots and additional rephrased medium- to high-quality documents.

\paragraph{High-quality web data.}
We include approximately 25B tokens from Nemotron-CC-v2.1-High-Quality. This subset incorporates recent Common Crawl snapshots and high-quality data translated into English from multiple languages. Additional LLM-based filtering removes uninformative translated documents. The inclusion of this data preserves exposure to natural web text during the annealing phase.

\paragraph{Mathematics data.}
Finally, we include approximately 126B tokens from Nemotron-CC-Math-v1, using only documents with quality scores of 4 and above. Nemotron-CC-Math-v1 is a high-quality math pre-training corpus built from Common Crawl using a pipeline designed to preserve equations and code, convert mathematical notation to standardized LaTeX, and remove noise. This component is included to strengthen mathematical reasoning and symbolic problem-solving capabilities.

\section{Post-Training}

\begin{figure}[H]
\centering
\resizebox{1.0\textwidth}{!}{%
\begin{tikzpicture}[
    stage/.style={
        draw=black,
        line width=0.5pt,
        fill=white,
        rounded corners=4pt,
        minimum width=2.55cm,
        minimum height=0.95cm,
        align=center,
        font=\small
    },
    milestone/.style={
        stage,
        line width=1.4pt,
        font=\small\bfseries
    },
    flowarrow/.style={-{Latex[length=2.0mm]}, line width=0.45pt}
]
\path[fill=gray!12, rounded corners=10pt]
    (-2.10,-2.90) rectangle (14.35,1.10);

\node[stage] (pre-train) at (0,0) {Pre-Training};
\node[stage] (anneal) at (3.05,0) {High-Quality\\Annealing};
\node[milestone] (base) at (6.10,0) {Loopie\\Base};
\node[stage] (spt) at (6.10,-1.70) {Supervised\\Pre-Training};
\node[stage] (mathrl) at (9.15,-1.70) {Math RL};
\node[milestone] (thinking) at (12.20,-1.70) {Loopie\\Thinking};

\draw[flowarrow] (pre-train.east) -- (anneal.west);
\draw[flowarrow] (anneal.east) -- (base.west);
\draw[flowarrow] (base.south) -- (spt.north);
\draw[flowarrow] (spt.east) -- (mathrl.west);
\draw[flowarrow] (mathrl.east) -- (thinking.west);
\end{tikzpicture}%
}
\caption{Overview of the Loopie training pipeline. The model is first pre-trained and annealed to produce \textbf{Loopie Base}, which then undergoes supervised pre-training and Math RL to produce \textbf{Loopie Thinking}.}
\label{fig:posttraining-pipeline}
\end{figure}
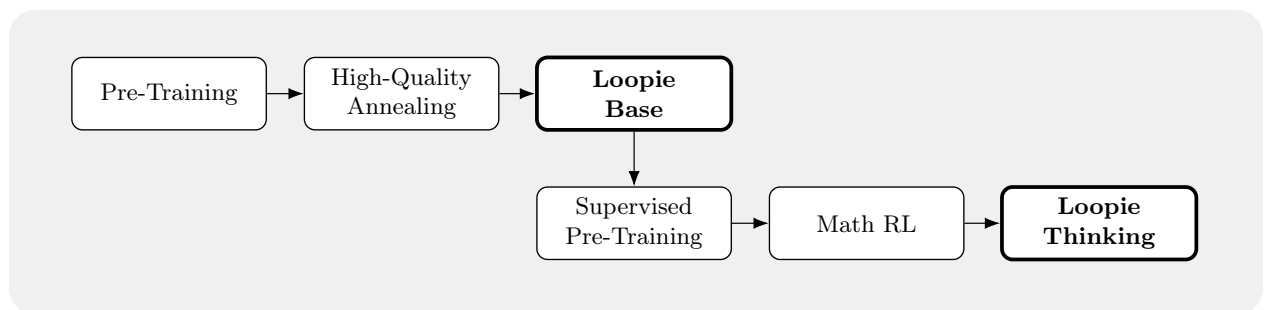

Loopie's post-training recipe consists of two main stages following Stage 2 high-quality annealing. First, we introduce a supervised pre-training stage in which we continue training the model on 2T tokens of instruction-following, reasoning, coding, mathematics, and tool-use data, allowing the base model to acquire broad task-following and problem-solving capabilities while preserving the knowledge and general capabilities learned during pre-training. Second, we apply large-scale reinforcement learning to further enhance Loopie's reasoning capabilities, improve its long-horizon problem solving, and align the model so that it produces reliable thinking traces. Together, these stages transform \textbf{Loopie Base} into the final \textbf{Loopie Thinking} model.

\subsection{Supervised Pre-Training}
\label{sec:spt}

We introduce \emph{supervised pre-training} (SPT), a training
regime that combines the supervision pattern of supervised fine-tuning
(SFT) with the optimization scale of language-model pre-training (PT).
SPT uses exactly the same training data and token-level objective as
SFT: prompt and context tokens are excluded from the loss, and only
supervised target tokens contribute to optimization. Unlike conventional
SFT, however, SPT uses global batch sizes, sequence lengths, and token
budgets typical of language-model pre-training.

\begin{table}[H]
\centering
\caption{
Comparison of supervised pre-training (SPT), conventional supervised
fine-tuning (SFT), and pre-training (PT). The SPT column reports the
representative configuration used for \textsc{Loopie}.
$\uparrow$, $\downarrow$, and $\approx$ indicate improvement,
degradation, and no material change, respectively, relative to the
corresponding starting checkpoint. Stability entries summarize our
observations over the evaluated training horizon rather than universal
guarantees.
}
\label{tab:spt-comparison}
\small
\setlength{\tabcolsep}{5pt}
\renewcommand{\arraystretch}{1.16}
\begin{tabularx}{
    \textwidth
}{
    @{}
    >{\raggedright\arraybackslash}p{0.29\textwidth}
    >{\centering\arraybackslash}X
    >{\centering\arraybackslash}X
    >{\centering\arraybackslash}X
    @{}
}
\toprule
&
\textbf{SPT}
&
\textbf{SFT}
&
\textbf{PT}
\\
\midrule

Loss function
&
Cross-entropy
&
Cross-entropy
&
Cross-entropy
\\

Loss-bearing positions
&
Supervised target tokens only
&
Supervised target tokens only
&
All non-padding tokens
\\

Pre-training metrics
&
$\uparrow$
&
$\downarrow$
&
$\uparrow$
\\

Reasoning metrics
&
$\uparrow$
&
$\uparrow$
&
$\approx$
\\

Overfitting after multiple epochs
&
Not observed
&
Observed
&
Not observed
\\

Global batch size
&
$\geq 1{,}024$
&
$32$--$128$
&
$\geq 1{,}024$
\\

Sequence length
&
$\geq 128\,\mathrm{K}$
&
$8$--$32\,\mathrm{K}$
&
$4$--$8\,\mathrm{K}$
\\

Nominal token positions per batch
&
$\geq 128\,\mathrm{M}$
&
$128\,\mathrm{K}$--$1\,\mathrm{M}$
&
$8$--$32\,\mathrm{M}$
\\

Total training token budget
&
$\geq 2\,\mathrm{T}$
&
$10$--$100\,\mathrm{B}$
&
$\geq 4\,\mathrm{T}$
\\

\bottomrule
\end{tabularx}

\vspace{2pt}
\end{table}

SPT separates two design choices that are commonly coupled:
(i) which tokens contribute to the training loss, and
(ii) the scale at which optimization is performed.
SFT and SPT share the former choice, whereas PT and SPT share the
latter. Consequently, SPT does not interpolate between an SFT loss and
a PT loss. Instead, it applies an SFT-style supervised objective in a
PT-scale optimization regime.

\paragraph{Training objective.}
Consider a supervised example
$z_i = [c_i; y_i]$, where $c_i$ denotes the input context and $y_i$
denotes the supervised target response. Let $w_{i,t} \in \{0,1\}$ be a
binary loss mask for the token at position $t$. All three training
regimes use token-level cross-entropy:
\begin{equation*}
\mathcal{L}_{\mathrm{CE}}(\theta; w)
=
-
\frac{
    \displaystyle
    \sum_{i=1}^{B}
    \sum_{t=1}^{T_i}
    w_{i,t}
    \log p_{\theta}
    \left(
        z_{i,t} \mid z_{i,<t}
    \right)
}{
    \displaystyle
    \sum_{i=1}^{B}
    \sum_{t=1}^{T_i}
    w_{i,t}
}.
\label{eq:spt-loss}
\end{equation*}
For both SPT and SFT, $w_{i,t}=1$ only when $z_{i,t}$ belongs to the
supervised target $y_i$; prompt, context, and padding positions are
masked out. In conventional PT, by contrast, $w_{i,t}=1$ for every non-padding token. Thus, all three regimes employ the same cross-entropy loss, but differ in the positions to which the loss is applied and in the scale of optimization.

Table~\ref{tab:spt-comparison} summarizes these distinctions. For representative SFT settings, including global batch sizes, training token counts, and sequence lengths, we primarily follow the configurations reported for OLMo 3, the Nemotron 3 series, and Step-3.5-Flash~\citep{olmo2026olmo3,nvidia2025nemotron3nanoopen,bercovich2025llamanemotronefficientreasoningmodels,nvidia2026nemotron3superopen,nvidia2026nemotron3ultraopen,huang2026step35flashopen}. We use
\emph{pre-training metrics} to refer to metrics that evaluate general base-model capabilities and \emph{reasoning metrics} to refer to metrics that primarily evaluate performance on challenging reasoning tasks, such as competition-level mathematics and coding problems. In our experiments, SPT improves performance on both types of metrics simultaneously. In contrast, the conventional SFT baseline only improves performance on reasoning metrics but degrades performance on pre-training metrics, whereas PT improves pre-training metrics performance while leaving performance on reasoning metrics approximately unchanged.

\begin{figure}[!t]
\centering
\includegraphics[width=0.8\textwidth]{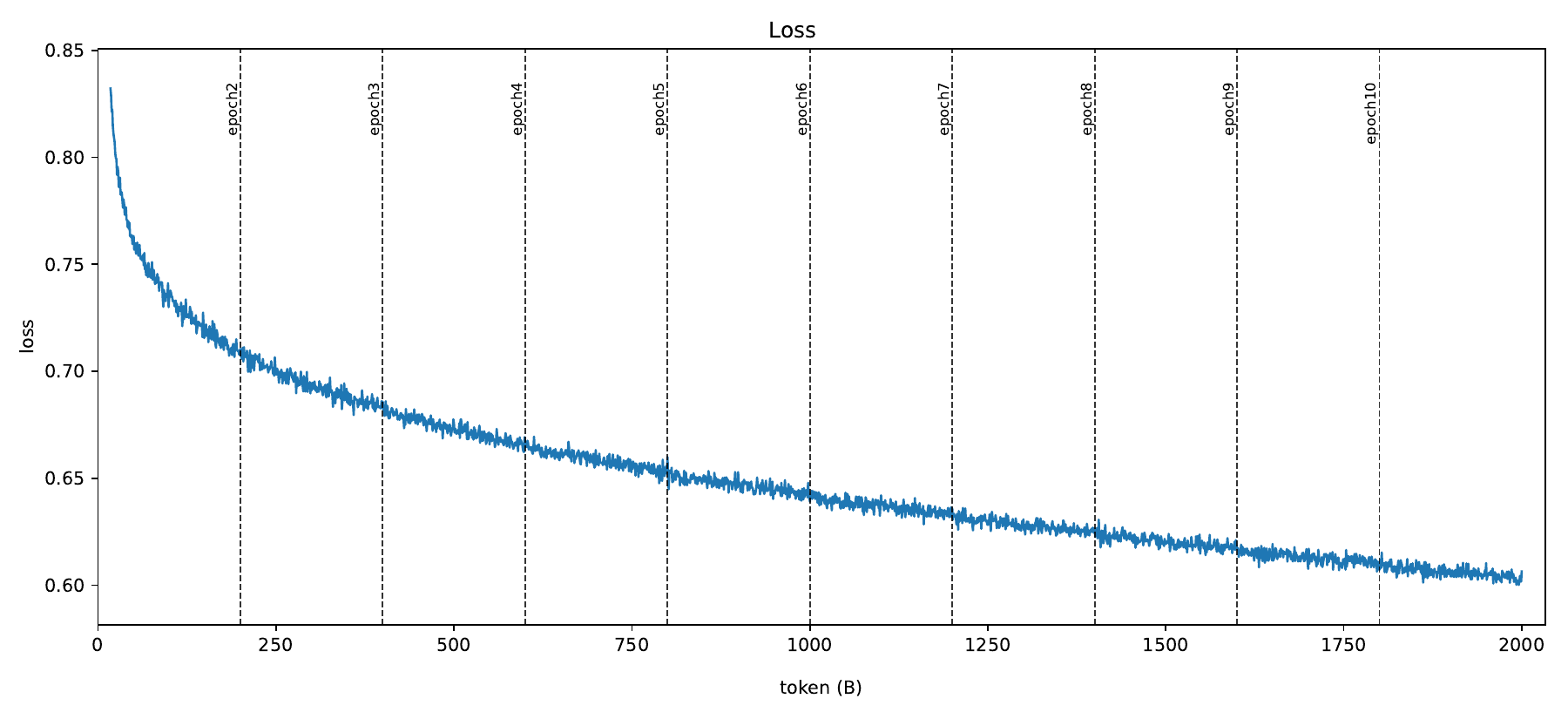}
\caption{Training loss for Loopie-6B-A0.6B during supervised pre-training. Because the evaluation and training loss curves nearly overlap, only the training loss curve is shown. Over 10 epochs of supervised pre-training on a total of 2 trillion tokens, the loss decreases smoothly, with no loss cliff observed at epoch boundaries.}
\end{figure}

\begin{figure}[H]
\centering
\includegraphics[width=1.08\textwidth]{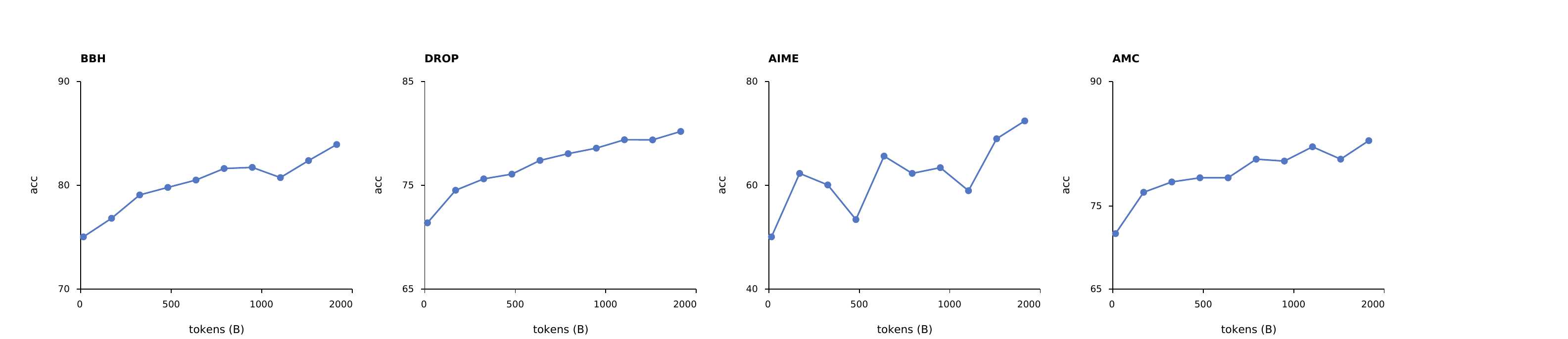}
\caption{Trends in reasoning metrics during supervised pre-training. Throughout training on 2 trillion tokens, performance on the reasoning metrics continues to improve, with no sign of slowing.}
\label{fig:spt-reasoning}
\end{figure}

\paragraph{Pre-training-scale optimization over supervised data.}
SPT processes 128 million tokens per global batch, approximately 1,000 times the number processed per batch in conventional SFT. This makes overfitting much less likely in SPT than in standard SFT. This scale gives SPT distinct training dynamics, consistent with observations that optimization hyperparameters and scaling behavior can change in large-language-model training regimes~\citep{jin2023rethinkinglearningratetuning}. Under SPT, the model can be trained smoothly for substantially more epochs without showing signs of overfitting, such as abrupt drops in loss at epoch boundaries. At the same time, performance on downstream metrics continues to improve steadily. In addition, SPT mitigates catastrophic forgetting~\citep{wu2024mitigating}, which is a major drawback of conventional SFT and typically causes substantial degradation of the general knowledge acquired during pre-training. Notably, SPT not only avoids this degradation but also further improves general-knowledge performance.

\begin{figure}[H]
\centering
\hspace*{-4mm}\includegraphics[width=0.85\textwidth]{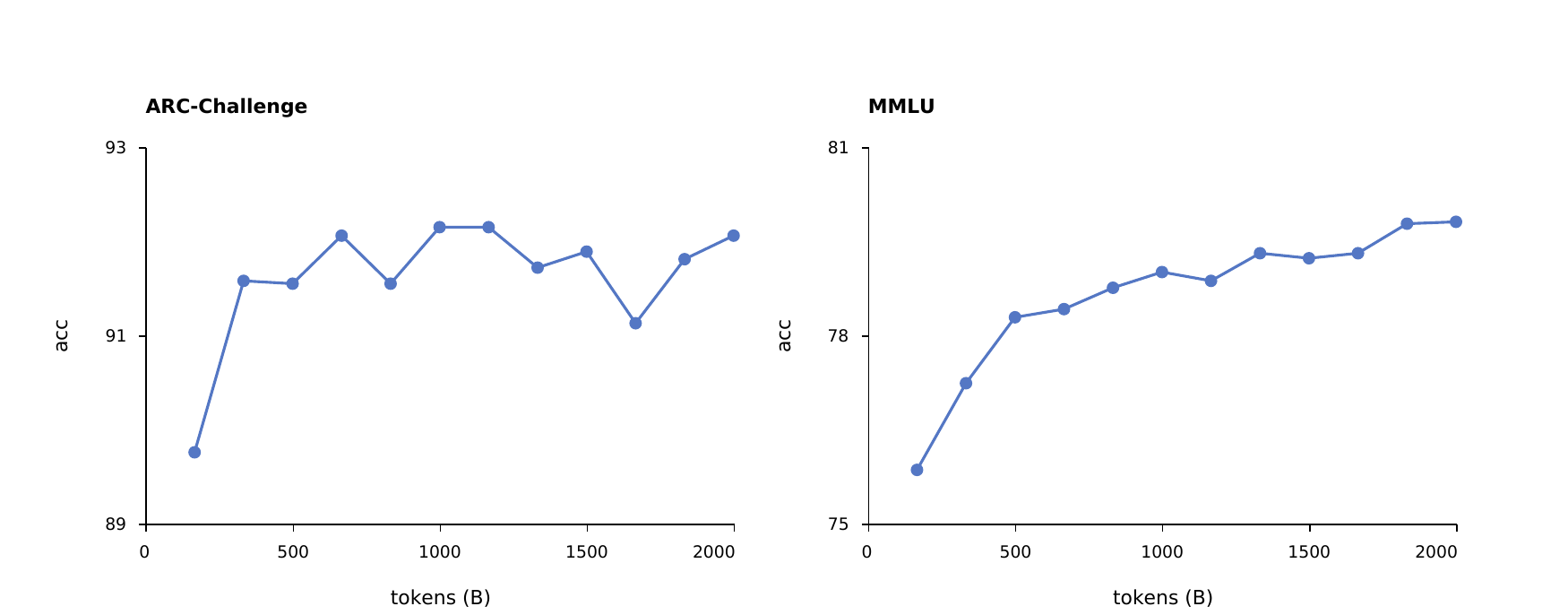}\hspace*{4mm}
\caption{Trends in pre-training metrics during supervised pre-training. ARC-Challenge represents general reasoning performance, while MMLU represents general knowledge performance. Throughout training on 2 trillion tokens, performance on the pre-training metrics does not degrade; instead, it continues to improve.}
\label{fig:spt-pt}
\end{figure}

As shown in Figure~\ref{fig:spt-pt}, SPT consistently improves performance on downstream pre-training metrics, including ARC-Challenge and MMLU, throughout approximately 10 epochs of training on 2T tokens. This finding challenges the conventional view that SFT leads to catastrophic forgetting.

Simply running a conventional SFT configuration for more epochs does not reproduce the optimization regime of SPT. Conventional-scale SFT often begins to overfit between the second and fourth epochs. In small-batch SFT, each epoch involves many parameter updates computed from small batches, and repeated exposure to the same examples can lead to rapid specialization and memorization. In contrast, every SPT update aggregates supervision from more than one thousand sequences and over one hundred million nominal token positions. We hypothesize that this broad gradient aggregation, together with long contexts and a large token budget, mitigates the overly narrow specialization commonly observed during repeated SFT.

\subsection{Reinforcement Learning}
\label{sec:rl}

After SPT, we apply a reinforcement-learning stage to obtain \textbf{Loopie Thinking}. Following prior work, we do not mix data from multiple domains during training because domain-specific length biases may interfere with one another~\citep{chen2026acereasonnemotron,liu2026acereasonnemotron}. We first conduct reinforcement learning on mathematical tasks; once performance saturates, we continue with reinforcement learning on coding tasks.

\paragraph{Algorithm.}
Our optimizer builds on Group Sequence Policy Optimization (GSPO) \citep{zheng2025groupsequencepolicyoptimization}, with the asymmetric clipping and dynamic sampling techniques introduced by DAPO \citep{yu2025dapo}. For each prompt $q$, we sample a group of $G$ completions $\{o_i\}_{i=1}^{G}$ from the old policy $\pi_{\theta_{\mathrm{old}}}$ and score each completion using a verifier, yielding outcome rewards $\{R_i\}_{i=1}^{G}$.
We estimate sequence-level advantages by normalizing rewards within the
sampled group:
\begin{equation*}
\hat{A}_i
=
\frac{
    R_i - \operatorname{mean}(\{R_j\}_{j=1}^{G})
}{
    \operatorname{std}(\{R_j\}_{j=1}^{G}) + \epsilon
},
\label{eq:rl-group-advantage}
\end{equation*}
where $\epsilon$ is a small numerical constant.

Unlike token-level policy optimization, GSPO defines a length-normalized
sequence-level importance ratio:
\begin{equation*}
s_i(\theta)
=
\left(
    \frac{\pi_{\theta}(o_i \mid q)}{\pi_{\theta_{\mathrm{old}}}(o_i \mid q)}
\right)^{\frac{1}{|o_i|}}
=
\exp
\left(
    \frac{1}{|o_i|}
    \sum_{t=1}^{|o_i|}
    \log
    \frac{\pi_{\theta}(o_{i,t} \mid q, o_{i,<t})}{\pi_{\theta_{\mathrm{old}}}(o_{i,t} \mid q, o_{i,<t})}
\right).
\end{equation*}
That is, $s_i(\theta)$ is the geometric mean of the token-level
importance ratios over the response. The length normalization reduces
the variance of the sequence-level ratio and keeps responses of different
lengths within a comparable numerical range.

The policy is optimized using a sequence-level clipped objective:
\begin{equation*}
\mathcal{J}_{\mathrm{RL}}^{\mathrm{GSPO}}(\theta)
=
\mathbb{E}_{
    q \sim \mathcal{D},\,
    \{o_i\}_{i=1}^{G}
    \sim \pi_{\theta_{\mathrm{old}}}(\cdot \mid q)
}
\left[
    \frac{1}{G}
    \sum_{i=1}^{G}
    \min
    \left(
        s_i(\theta)\hat{A}_i,\,
        \operatorname{clip}
        \left(
            s_i(\theta),
            1-\epsilon_{\mathrm{low}},
            1+\epsilon_{\mathrm{high}}
        \right)
        \hat{A}_i
    \right)
\right].
\label{eq:gspo-objective}
\end{equation*}
In contrast to token-level clipping, GSPO applies the clipping operation
to the entire response. Consequently, all tokens in the same response
share the same sequence-level importance weight, aligning the unit of
off-policy correction and optimization with the unit of reward.

Following the Clip-Higher strategy, we use
$\epsilon_{\mathrm{high}} > \epsilon_{\mathrm{low}}$.
Under GSPO, this asymmetric clipping is applied to the sequence-level importance ratio rather than individual token-level ratios. The larger upper clipping range allows positively advantaged exploratory responses
to receive stronger probability-increasing updates, while the more
conservative lower clipping range limits excessive probability decreases
for negatively advantaged responses.

We further apply prompt-level dynamic filtering. A prompt is retained only
when its sampled group contains both successful and unsuccessful completions:
\begin{equation*}
0
<
\sum_{i=1}^{G}
\mathbf{1}\{R_i = R_{\mathrm{pass}}\}
<
G.
\label{eq:dynamic-filtering}
\end{equation*}
Groups that are entirely correct or entirely incorrect provide no useful
relative preference signal for GRPO. We therefore oversample candidate
prompts and filter out such groups until the effective batch is full. This
keeps the optimization focused on prompts that yield non-degenerate policy
gradients.

\paragraph{Training data.}
We use the mathematics and code subsets of the Guru-RL corpus
\citep{cheng2025guru}. The mathematics pool is drawn from the OR1, DAPO, and
DeepScaleR sources \citep{he2025skywork,yu2025dapo,tan2025deepscaler}. The
code pool combines LeetCodeDataset, TACO-Verified, PrimeIntellect/SYNTHETIC-1,
and historical LiveCodeBench training problems
\citep{xia2025leetcodedataset,li2023taco,li2024tacoverified,
mattern2025synthetic1,jain2024livecodebench}. We apply an additional
curation pass that removes malformed prompts, unverifiable answers, flaky
tests, duplicate examples, and examples that substantially overlap with
held-out evaluation sets. Rewards for mathematical problems are computed using
rule-based answer equivalence, whereas rewards for coding problems are computed
using sandboxed unit-test execution.

\paragraph{Training schedule.}
We train in two context-length stages. The first stage uses a maximum response
length of $32$K tokens. This stage is computationally efficient because most
rollouts terminate naturally within this limit. When the rollout truncation
rate exceeds $10\%$, we
switch to a $64$K-token stage, giving the policy additional room for longer
derivations and code-reasoning trajectories. We continue RL until the aggregate
validation score stops improving and begins to decline; the checkpoint
immediately before sustained degradation is selected as \textbf{Loopie
Thinking}. Throughout training, we monitor validation accuracy, mean response
length, generation entropy, and truncation rate to detect late-stage
over-optimization.







\subsection{Results}

We report two complementary comparisons of the final Loopie Thinking models. We compare Loopie-20B-A2B with MoE reasoning models at a similar active-parameter scale, and Loopie-6B-A0.6B with a broader set of compact reasoning models. All evaluations in this section are conducted using the EvalScope framework, and the IFEval score is reported under the \texttt{inst\_level\_loose} setting. For AIME 2024 and AIME 2025, we report avg$@$8 results; for all other benchmarks, we report pass$@$1. The AIME result shown in the teaser figure is for AIME 2024.

\begin{table}[H]
\centering
\caption{Comparison of Loopie-20B-A2B with similarly sized MoE reasoning models across knowledge, general, code, and math benchmarks.}
\label{tab:model-comparison}
\small
\setlength{\tabcolsep}{0.1pt}
\newcommand{\modelheader}[2]{\makecell{\scalebox{0.84}{\textbf{#1}}\\\scalebox{0.84}{\textbf{#2}}}}
\newcommand{\modelheaderthree}[3]{\makecell{\scalebox{0.84}{\textbf{#1}}\\\scalebox{0.84}{\textbf{#2}}\\\scalebox{0.84}{\textbf{#3}}}}
\makebox[\linewidth][c]{%
\scalebox{1.1}{%
\begin{adjustbox}{max width=\linewidth}
\begin{tabular}{@{}>{\hspace{2pt}\raggedright\arraybackslash}p{3.2cm}|
>{\centering\arraybackslash}p{2cm}@{\hspace{-0.144cm}}
>{\centering\arraybackslash}p{2cm}@{\hspace{-0.144cm}}
>{\centering\arraybackslash}p{2cm}@{\hspace{-0.144cm}}
>{\centering\arraybackslash}p{2cm}@{\hspace{-0.144cm}}
>{\centering\arraybackslash}p{2cm}@{}}
\hline
\textbf{} &
\modelheaderthree{Qwen3}{30B-A3B}{Thinking} &
\modelheaderthree{Nemotron}{3 Nano}{30B-A3B} &
\modelheaderthree{Nemotron}{Cascade 2}{30B-A3B} &
\modelheaderthree{GPT-OSS}{20B-A2B}{High} &
\modelheaderthree{Loopie}{20B-A2B}{Thinking} \\

\hline
\rule[-1\baselineskip]{0pt}{2.5\baselineskip}Pre-training tokens & 36T & 25T & 25T & Unknown & \textbf{3.5T} \\
\hline
\textbf{Knowledge} & & & & & \\
MMLU & 85.83 & 80.52 & 81.22 & 81.64 & 81.28 \\
MMLU-Redux & 88.25 & 82.54 & 83.89 & 83.40 & 83.61 \\
\hline
\textbf{General} & & & & & \\
ARC-Challenge & 96.67 & 91.96 & 93.86 & 92.42 & 93.52 \\
DROP & 87.70 & 85.92 & 79.06 & 62.82 & 82.08 \\
BBH & 86.59 & 68.76 & 75.86 & 84.76 & 82.28 \\
SciQ & 95.40 & 93.40 & 92.40 & 93.80 & 92.50 \\
IFEval & 85.58 & 77.05 & 79.21 & 70.03 & 84.72 \\
\hline
\textbf{Code} & & & & & \\
MBPP & 96.50 & 90.66 & 80.54 & 78.99 & 89.49 \\
MBPP+ & 98.68 & 91.53 & 83.60 & 97.62 & 83.07 \\
HumanEval & 96.95 & 86.59 & 79.88 & 92.07 & 89.02 \\
HumanEval+ & 92.07 & 87.80 & 78.66 & 89.02 & 87.20 \\
\hline
\textbf{Math} & & & & & \\
AIME 24 & 90.10 & 85.00 & 93.33 & 88.33 & 92.09 \\
AIME 25 & 83.75 & 74.17 & 91.25 & 87.50 & 83.75 \\
AMC & 96.73 & 91.80 & 93.57 & 91.05 & 94.21 \\
OlympiadBench & 81.20 & 76.68 & 88.64 & 70.03 & 80.50 \\
\hline
\end{tabular}
\end{adjustbox}%
}%
}%
\end{table}

Table~\ref{tab:model-comparison} highlights the pre-training efficiency of Loopie-20B-A2B. Loopie is pre-trained on only 3.5T tokens, whereas Nemotron 3 Nano and Nemotron Cascade 2 are each trained on 25T tokens drawn from the same pre-training data source. Thus, Loopie uses less than one seventh of their pre-training tokens, yet matches or outperforms both models on most knowledge and general-capability benchmarks. In particular, Loopie reaches 81.28 on MMLU, exceeding both Nemotron 3 Nano (80.52) and Nemotron Cascade 2 (81.22). On ARC-Challenge, Loopie scores 93.52, outperforming Nemotron 3 Nano by 1.56 points and coming within 0.34 points of Nemotron Cascade 2. It also substantially surpasses both models on BBH, with a score of 82.28 versus 68.76 and 75.86, and on IFEval, with 84.72 versus 77.05 and 79.21.

This advantage extends across the broader pre-training evaluation suite: Loopie exceeds at least one of the two Nemotron models on every reported knowledge or general-capability benchmark and outperforms each of them on five of the seven benchmarks, despite having a pre-training token budget more than seven times smaller. The resulting profile is therefore not merely competitive at a smaller training budget; it indicates substantially higher token efficiency under the same pre-training data. Loopie also retains strong post-training reasoning performance, reaching 92.09 on AIME 2024 and 94.21 on AMC, ranking second among the listed models on both benchmarks, while its AIME 2025 score of 83.75 ties Qwen3-30B-A3B Thinking.

\begin{table}[H]
\centering
\small
\setlength{\tabcolsep}{0.1pt}
\newcommand{\modelheader}[2]{\makecell{\scalebox{0.84}{\textbf{#1}}\\\scalebox{0.84}{\textbf{#2}}}}
\newcommand{\modelheaderthree}[3]{\makecell{\scalebox{0.84}{\textbf{#1}}\\\scalebox{0.84}{\textbf{#2}}\\\scalebox{0.84}{\textbf{#3}}}}
\makebox[\linewidth][c]{%
\scalebox{1.1}{%
\begin{adjustbox}{max width=\linewidth}
\begin{tabular}{@{}>{\hspace{2pt}\raggedright\arraybackslash}p{3.2cm}|
>{\centering\arraybackslash}p{2cm}@{\hspace{-0.144cm}}
>{\centering\arraybackslash}p{2cm}@{\hspace{-0.144cm}}
>{\centering\arraybackslash}p{2cm}@{\hspace{-0.144cm}}
>{\centering\arraybackslash}p{2cm}@{\hspace{-0.144cm}}
>{\centering\arraybackslash}p{2cm}@{\hspace{-0.144cm}}
>{\centering\arraybackslash}p{2cm}@{\hspace{-0.144cm}}
>{\centering\arraybackslash}p{2cm}@{\hspace{-0.144cm}}
>{\centering\arraybackslash}p{2cm}@{}}
\hline
\textbf{} &
\modelheaderthree{DeepSeek-R1}{Distill-Qwen}{1.5B} &
\modelheader{Gemma-4}{E2B-it} &
\modelheader{Gemma-4}{E4B-it} &
\modelheader{Qwen3}{1.7B} &
\modelheader{MiniCPM5}{1B} &
\modelheader{Ouro 1.4B}{Thinking} &
\modelheader{Ouro 2.6B}{Thinking} &
\modelheader{Loopie}{6B-A0.6B} \\

\hline
\textbf{Knowledge} & & & & & & & & \\
MMLU & 44.85 & 63.33 & 72.97 & 69.08 & 61.54 & 72.40 & 82.70 & 78.36 \\
MMLU-Redux & 51.28 & 72.28 & 79.61 & 74.28 & 70.63 & 73.75 & 86.28 & 81.35 \\
\hline
\textbf{General} & & & & & & & & \\
ARC-Easy & 69.19 & 87.08 & 92.97 & 91.20 & 85.27 & 90.87 & 94.28 & 93.77 \\
ARC-Challenge & 61.09 & 83.28 & 90.02 & 86.09 & 75.34 & 88.48 & 93.94 & 91.13 \\
SciQ & 44.10 & 86.80 & 94.40 & 87.70 & 82.48 & 91.60 & 94.90 & 90.10 \\
\hline
\textbf{Code} & & & & & & & & \\
MBPP & 35.80 & 77.04 & 82.49 & 80.54 & 50.97 & 88.63 & 95.69 & 76.65 \\
MBPP+ & 63.76 & 86.77 & 90.21 & 82.80 & 81.75 & 89.89 & 96.01 & 84.66 \\
HumanEval & 65.85 & 79.27 & 91.46 & 85.98 & 86.59 & 95.12 & 95.12 & 84.15 \\
HumanEval+ & 65.24 & 81.71 & 85.98 & 82.93 & 86.59 & 87.80 & 89.02 & 79.88 \\
\hline
\textbf{Math} & & & & & & & & \\
AIME 24 & 35.83 & 38.33 & 49.17 & 49.58 & 47.50 & 50.83 & 62.50 & 80.42 \\
AIME 25 & 24.17 & 26.25 & 35.42 & 35.00 & 38.75 & 44.17 & 51.67 & 70.83 \\
AMC & 67.91 & 75.37 & 85.08 & 73.88 & 28.36 & 78.36 & 85.82 & 84.33 \\
MATH-500 & 83.60 & 89.00 & 91.60 & 90.60 & 60.00 & 92.20 & 92.20 & 93.80 \\
GSM8K & 72.18 & 90.60 & 93.40 & 90.14 & 76.35 & 94.09 & 96.13 & 93.63 \\
\hline
\end{tabular}
\end{adjustbox}%
}%
}%

\vspace{2pt}
\end{table}

\section{Related Work}
\label{sec:related_work}

\subsection{Looped Transformers}

We use the term \emph{looped model} to denote an architecture that reuses a learned internal operator within a single forward computation, rather than repeatedly calling an otherwise complete model. This design lineage predates contemporary language models: Neural GPUs repeatedly apply a shared convolutional transition to learn algorithms, Adaptive Computation Time (ACT) learns how many recurrent updates to execute, Universal Transformers (UTs) tie self-attention and transition blocks across depth, and Deep Equilibrium Models (DEQs) avoid explicit unrolling by solving for a fixed point of a weight-tied transformation \citep{kaiser2016neural,graves2016adaptive,dehghani2018universal,bai2019deep,bai2020multiscale}. Work on training implicit models and on generalist neural algorithmic learners further established practical optimization methods and reusable processors for effectively unbounded or task-shared computation. These are direct architectural antecedents of looped Transformers, not merely generic recurrent or test-time-compute methods.

Across Transformer families, layer tying, recurrent state, adaptive depth, and memory reuse also appear in ALBERT, tied Transformers, depth-adaptive Transformers, feedback memory, recurrent-memory Transformers, and block-recurrent Transformers \citep{lan2019albert,xia2019tied,elbayad2020depth,fan2020feedback,bulatov2022rmt,hutchins2022blockrecurrent}. Modern variants make the shared-depth computation itself the central scaling axis. MoEUT recurrently reuses fine-grained expert groups, Relaxed Recursive Transformers supplement a repeated block with depth-specific low-rank adapters, and implicit state-space language models iterate a shared transition toward a fixed point while preserving substantial training parallelism \citep{csordas2024moeut}. Thus, rather than relying on ordinary depth scaling, looped Transformers learn a reusable transition $h_{t+1}=F_{\theta}(h_t,x,e_t)$, where $e_t$ may encode the loop index, a halting state, or an injected copy of the input. Effective inference depth can then vary without introducing an independently parameterized layer at every step.

\subsection{Inductive Biases}

Looped depth imposes several biases that are weak or absent in a stack of independent Transformer layers. \emph{First, recurrent operator sharing} encourages a reusable update rule rather than a sequence of layer-specific feature maps. This is the defining recursive bias of UTs and tied-depth models, and theoretical analyses connect it to fixed-point iteration, gradient-based in-context learning, normalized gradient descent, and power iteration \citep{dehghani2018universal,takase2023lessons,yang2023looped,gatmiry2024can,chen2025bypassing,wu2026powermethod}. \emph{Second, shared depth decouples effective computation from parameter count}: a compact recurrent core can be unrolled for more steps, favoring iterative refinement over memorizing a separate computation in each layer \citep{lan2019albert,saunshi2025latentthoughts,geiping2025scalinglatent,zhu2025scalinglatent,schwethelm2026much}.

\emph{Third, looping induces an iterative-algorithm bias}. When a target computation is naturally expressed by repeated application of a small rule set, looped models can emulate programs, optimization procedures, graph algorithms, and structured algorithmic processors, often improving length or depth extrapolation when the loop count adapts to problem size \citep{giannou2023looped,backdeluca2024simulation,gao2025algoformer,fan2024looped,xu2025cotloop}. \emph{Fourth, weight sharing creates a knowledge re-access and compositionality bias}: after each new bridge entity is inferred, the same retrieval, binding, and update machinery can be applied again. This directly targets failures of implicit multi-hop composition in ordinary Transformers and motivates architectures that carry discrete and continuous states or align latent and explicit reasoning \citep{wang2024grokking,biran2024hopping,guo2025twohop,yao2025implicit,kohli2026loopthinkgeneralize,fu2026discoloop,fan2026lotus}. \emph{Fifth, recurrence provides a latent scratchpad and an adaptive-computation bias}. Additional computation occurs in the hidden state rather than through the generation of extra verbal tokens, while timestep conditioning, token-wise routing, dynamic halting, fixed-point convergence, and shortcut consistency allocate more computation to harder inputs \citep{saunshi2025latentthoughts,geiping2025scalinglatent,bay2025mixture,fu2025thinkathard,jeddi2026loopformer,movahedi2026fixedpoint}. These benefits are not automatic: looped models can drift, overthink, or collapse to shallow computation unless their state injection, normalization, supervision, and stopping rules are designed for stable iteration \citep{chowdhury2024investigating,labovich2026stability,rauba2026tarm,moosa2026dynamiccompute}.

\subsection{Theory, Mechanisms, and Scaling}

Formal results characterize both the capabilities and the limits of shared-depth computation. Looped Transformers can execute instruction-level programs, simulate latent chains of thought, and achieve better approximation rates through timestep modulation \citep{giannou2023looped,saunshi2025latentthoughts,xu2025expressive}. More specialized analyses show that a recurrently reused attention layer can learn normalized-gradient updates for in-context logistic regression or implement the power method under layer normalization \citep{wu2026powermethod}. At the same time, compressed recurrent states create a memory-budget separation relative to full sequence-state chain-of-thought methods, and set-valued or graph computations require careful distinctions among fixed points, convergence, and halting. A complementary reinforcement-learning analysis formalizes how additional internal computation changes the class of compute-bounded policies.

A second line of work studies optimization, stability, and scaling with loop count. Fixed-point analyses identify conditions under which recurrent states remain reachable, input-dependent, and trainable; residual-scaling theory argues that correlated tied-block updates require loop-aware scaling; and iso-depth studies quantify the exchange rate among recurrent passes, unique depth, and training compute \citep{labovich2026stability,wang2026residualscaling,schwethelm2026much}. Two-scale latent-dynamics analyses derive convergence-sensitive early-exit criteria, while controlled studies of adaptive computation show that learned loop allocation can correlate with token difficulty without necessarily extrapolating to longer inputs \citep{pappone2025twoscale,moosa2026dynamiccompute}. Memory tokens and ACT initialization can determine whether a UT enters a non-trivial reasoning regime, and hierarchical and flat recurrence can exhibit materially different optimization behavior even under shared-weight controls \citep{sapunov2026utm,han2026hierarchicalflat}. Compression studies likewise show that preserving per-cell accuracy or local states is insufficient if quantization or pruning damages the recursive trajectory needed for exact solutions.

Mechanistic work examines what is represented across recurrent steps. Probing studies find that recurrent language models often approach loop-specific fixed points or replay feed-forward-like inference stages, while evidence for a literal latent chain-of-thought is mixed and depth-dependent \citep{blayney2026mechanistic,lu2025latentcot}. Shared recurrent modules can nevertheless specialize to perform distinct functional roles through asymmetric state identities, and interaction-locality measurements reveal how repeated local updates accumulate into global puzzle structure. Other probes recover relational preference information from differences between loop states, and analyses of tabular foundation models use the observed layer redundancy to motivate a single repeatedly applied layer. Finally, recurrent-depth models can unlock systematic multi-hop generalization, but autoregressive controls show that the benefit depends on where compute is placed rather than on recursion alone \citep{kohli2026loopthinkgeneralize,rauba2026tarm}.

\subsection{Architectures, Training Objectives, and Inference}

Recent looped language models treat recurrence as a third scaling axis alongside parameter count and token count. Huginn and Ouro pretrain recurrent-depth language models whose shared cores can be unrolled for variable test-time depth, while retrofitted recurrence and LoopUS convert pretrained feed-forward models into latent-refinement systems without training a recurrent model entirely from scratch \citep{geiping2025scalinglatent,zhu2025scalinglatent,mcleish2025retrofitted,park2026loopus}. HRM-Text and CHERRY explore recurrent compression of deep language models, Hyperloop repeats only a middle block with cross-loop connections, and LoopMoE, sparse looped layers, and universal expert pools combine iterative depth with conditional capacity \citep{wang2026hrmtext,kwon2026cherry,zeitoun2026hyperloop,chen2026loopmoe,lee2026sparse}. MoEUT and Relaxed Recursive Transformers provide earlier expert-routing and low-rank mechanisms for recovering expressivity under parameter sharing \citep{csordas2024moeut}.

Dynamic-depth architectures decide not only \emph{how} to update a state but also \emph{where} and \emph{for how long} to recur. Mixture-of-Recursions routes tokens to different depths; CoTFormer exposes earlier recurrent representations and learns a compute-budgeted router; AdaPonderLM and Think-at-Hard allocate extra latent iterations selectively; and Chain-of-Layers methods skip or repeat pretrained layers at test time \citep{bay2025mixture,mohtashami2025cotformer,fu2025thinkathard,li2026programlayers}. LoopFormer regularizes trajectories across sampled depths, adaptive-loop models couple halting with external memory, and ChainGPT, MoDr, and depth-recurrent attention mixtures enrich the recurrent state transition through multi-rank updates, branch routing, or mixtures of sequence and depth attention \citep{jeddi2026loopformer,frey2026adaptive,zheng2026chaingpt,zhang2026modr}. SpiralFormer instead changes the resolution schedule across repeated applications, and subgoal-persistence models study when a hierarchical reasoner should re-plan rather than update the same latent plan at every step \citep{yu2026spiralformer,chadha2026replan}.

A parallel line of work targets stable and efficient unrolling. Parallel Loop Transformers share representations across loops, and LT2 replaces quadratic attention with a linear-time looped design \citep{wu2025parallellooptransformerefficient,deng2026lt2}. MELT keeps cache memory independent of loop depth; LASER compresses recursive activations during training; and Hyperloop and CHERRY reduce the number of distinct parameter blocks \citep{cakar2026laser,zeitoun2026hyperloop,kwon2026cherry}. Parcae constrains recurrent dynamics and derives scaling laws, CART anchors each iteration to context, fully looped signal routing stabilizes high loop counts, stochastic stopping improves extrapolation across unseen depths, and stability-aware recurrent training reduces drift under test-time scaling \citep{prairie2026parcae,capps2026cart,fu2026fullylooped,kuo2026stochasticstopping}. Training-free looping explores whether a pretrained model can be recurrently reused without architectural retraining \citep{chen2026trainingfree}.

Several methods supervise the \emph{trajectory} rather than the final answer alone. LoopRPT applies reinforcement pre-training across latent iterations, RLTT distributes reward over the latent trajectory, denoising recursion trains models to repeatedly correct corrupted targets, and Generative Recursive Reasoning expands recurrence into stochastic multi-trajectory generation \citep{tang2026looprpt,cameron2026denoisingrecursion,baek2026gram}. Probabilistic TRM injects noise at inference and selects among recursive trajectories, whereas LoopFormer uses shortcut consistency to align different computation budgets \citep{sghaier2026ptrm,jeddi2026loopformer}. Looped diffusion language models, fixed-point masked generative models, equilibrium reasoners, and attractor models connect explicit recurrence to denoising or convergence-based generation \citep{miele2026fixedpointmasked,feinashley2026attractor}. These approaches make intermediate-state quality, convergence geometry, and trajectory diversity first-class training targets rather than incidental by-products of depth.

\subsection{In-Context, Algorithmic, and Compositional Generalization}

In-context learning can be interpreted as executing an implicit learning algorithm over demonstrations. Bayesian, gradient-descent, preconditioned-gradient, linear-model, and causal-structure accounts motivate studying whether each recurrent step implements one additional internal update \citep{xie2022explanation,akyurek2023learning,vonoswald2023transformers,ahn2023transformers,li2023algorithms,zhang2023trained,nichani2024causal}. Looped Transformers make this correspondence explicit: they learn data-fitting algorithms with substantially fewer unique parameters, implement multi-step gradient procedures, and can use distinct preprocessing, looping, and postprocessing stages \citep{yang2023looped,gatmiry2024can,chen2025bypassing,gao2025algoformer}. Program-simulation results and normalized-gradient analyses further connect shared depth to reusable algorithmic primitives \citep{giannou2023looped}.

The same prior is relevant to length and compositional extrapolation. Neural GPUs and generalist recurrent processors learn repeated algorithmic updates, while looped Transformers simulate graph algorithms and obtain strong length generalization when the number of recurrent steps grows with the instance \citep{kaiser2016neural,backdeluca2024simulation,fan2024looped}. Timestep modulation, multi-resolution recursion, and depth-recurrent compositional models address the failure of a single stationary update to remain useful far beyond the trained depth \citep{xu2025expressive,yu2026spiralformer,chen2026thinking}. DiscoLoop carries both discrete embeddings and continuous hidden states across hops, and LOTUS aligns recurrent latent blocks with explicit chain-of-thought computation \citep{fu2026discoloop,fan2026lotus}. These results complement evidence that ordinary Transformers often memorize local facts or short computations yet fail to compose them systematically out of distribution \citep{biran2024hopping,yao2025implicit,dziri2023faith}.

\subsection{Abstract Reasoning}

ARC-style tasks are a natural stress test because they require inducing a new transformation from a few demonstrations rather than recalling a fixed skill \citep{chollet2019measure,chollet2025arcagi2}. HRM obtains substantial effective depth through coupled high- and low-level recurrent modules, TRM reduces the design to a tiny repeatedly applied network, and the URM study attributes much of the gain to Universal-Transformer-style recurrence and nonlinear depth computation \citep{wang2025hierarchical,jolicoeurmartineau2025trm,gao2025universal}. Controlled analyses caution that hierarchy, augmentation, identity conditioning, majority voting, and competition-time adaptation can materially affect reported scores \citep{ge2025hrmperspectives,ren2026reasoningguessing,royeazar2025trm,mcgovern2025testtime}.

Follow-up designs explore complementary ways to improve recursive abstract reasoning. CosmicFish-HRM adapts hierarchical recurrence to compact language models; Recursive Inference Machines generalize generator--solver recursion; and Fixed-Point Reasoners halt upon convergence rather than at a preset depth \citep{lakkapragada2026cosmicfish,komisarczyk2026rim,movahedi2026fixedpoint}. Generative, probabilistic, and denoising recursive models introduce trajectory diversity or iterative corruption-and-repair curricula \citep{baek2026gram,sghaier2026ptrm,cameron2026denoisingrecursion}. Equilibrium and attractor reasoners learn convergent solution dynamics, whereas Tiny Autoregressive Recursive Models test whether comparable compute is better spent on recurrence or ordinary autoregressive depth \citep{feinashley2026attractor,rauba2026tarm}. Memory-augmented UTs, LoopViT, and interaction-locality analyses expose complementary depth--state, visual-recurrence, and mechanism-level views \citep{sapunov2026utm,shu2026loopvit}. Overall, this literature supports recurrence as an architectural prior for iterative abstraction, while showing that successful test-time scaling depends on stable state transitions and informative intermediate supervision rather than loop count alone \citep{chollet2026arcprize,vahdati2026arcprogress}.

\section{Future Work}
Due to computational constraints, our post-training experiments focus primarily on mathematical and code reasoning tasks. We do not extensively explore other important capabilities studied in prior work, such as scientific question answering, instruction following, alignment with human conversational preferences, or agentic task-solving. A natural direction for future work is to continue to post-train Loopie on a broader set of tasks and preference signals, which may further improve its practical utility and general-purpose capabilities. Also, due to limited computational resources, we were unable to conduct a sufficiently comprehensive ablation study of supervised pre-training, which we leave for future work.

In addition, our study focuses primarily on matching compute budgets during pre-training, and we have not yet conducted systematic studies of inference-time computation. Matching and optimizing inference-time compute remain important directions for future work; prior work such as the Parallel Loop Transformer~\citep{wu2025parallellooptransformerefficient} provides a promising example of this approach.

Finally, our study intentionally focuses on applying looped Transformers to a relatively clean and well-controlled base architecture, Qwen3-30B-A3B, in order to avoid confounding factors introduced by architectural variations. As a result, we do not investigate several recent architectural advances that may be complementary to our method. Exploring how Loopie interacts with these newer designs remains an important direction for future work.

\section{Conclusion}
We introduced Loopie, a family of looped MoE language models that makes recurrent depth competitive under a matched pre-training compute budget. By combining layer-loop recurrence with a hardware-aware scaling recipe, Loopie consistently outperforms compute-matched vanilla Transformer baselines across model scales. A large-scale post-training pipeline based on Supervised Pre-training and reinforcement learning further equips Loopie with strong mathematical reasoning and coding abilities. These results suggest that recurrent computation, when jointly optimized with architecture and training efficiency, can serve as a practical scaling axis for large language models.

\newpage
\section*{Author Contributions}
\label{sec:contrib}

Zitian Gao completed model training, experimental implementation, and writing of this paper. The model architecture was designed by Zitian Gao, Yilong Chen, Yihao Xiao, and Xinyu Yang, with guidance from Ran Tao, Joey Zhou, and Bryan Dai. Yilong Chen and Xinyu Yang completed this work during their internship at IQuest Research.
\section{Acknowledgments}
\begin{itemize}
    \item We thank Benhao Huang at Carnegie Mellon University and Shaowen Wang at Tsinghua University for their valuable feedback and careful review.
    
    \item We thank Zhengmao Ye at IQuest Research for his infrastructure support.
    
    \item We thank NVIDIA Nemotron for its outstanding contributions to the open-source community. In particular, Loopies would not have achieved its current level of performance without the training data provided by Nemotron-CC-v2 and Nemotron-Cascade-2.
    
    \item We thank the Allen Institute for AI (AI2) for the elegant \LaTeX{} template used in this work.
\end{itemize}

\bibliographystyle{abbrvnat}
\bibliography{neurips_2023}

\clearpage
\appendix

\phantomsection
\addcontentsline{toc}{section}{Appendix}
\newcommand{\appsection}[1]{%
  \refstepcounter{section}%
  \section*{\thesection\quad #1}%
  \addcontentsline{toc}{subsection}{\protect\numberline{\thesection}#1}%
}

\appsection{Architecture Details}
\label{arch_details}
\begin{table}[!hbp]
\centering
\caption{Architecture Details of Loopie-20B-A2B and Loopie-6B-A0.6B}
\vspace{2pt}
{\small
\setlength{\tabcolsep}{20pt}
\renewcommand{\arraystretch}{1.5}
\begin{tabular}{l|cc}
\toprule
 & Loopie-20B-A2B & Loopie-6B-A0.6B  \\
\midrule
Layer-loop times & 2 & 2  \\
QK layer norm & Enabled & Enabled \\
Num layers & 27 & 18 \\
Hidden size & 2304 & 1536 \\
MoE hidden size & 832 & 576 \\
Attention type & GQA & GQA \\
Attention heads & 72 & 48 \\
Attention groups & 36 & 24 \\
Position embedding type & RoPE & RoPE \\
Rotary base & 10000 & 10000 \\
RMSNorm eps & 1e-6 & 1e-6 \\
MoE router topk & 8 & 8 \\
Num experts & 128 & 128 \\
MLP type & SwiGLU & SwiGLU \\
Tokenizer & Qwen3 & Qwen3 \\
Vocabulary size & 151936 & 151936 \\
\bottomrule
\end{tabular}
}
\label{tab:arch_details}
\end{table}

\newpage
\appsection{Pre-training Details}
\begin{table}[!hbp]
\centering
\caption{Pre-training Stage-1 Details of Loopie-20B-A2B and Loopie-6B-A0.6B}
\vspace{2pt}
{\small
\setlength{\tabcolsep}{20pt}
\renewcommand{\arraystretch}{1.5}
\begin{tabular}{l|cc}
\toprule
 & Loopie-20B-A2B & Loopie-6B-A0.6B  \\
\midrule
Initialization std & $1/\sqrt{2.5*2304}$ & $1/\sqrt{2.5*1536}$ \\
LM head Initialization std & $1/\sqrt{2304}$ & $1/\sqrt{1536}$ \\
Embedding Initialization std & $1/\sqrt{2304}$ & $1/\sqrt{1536}$ \\
MoE Auxiliary loss coeffient & 0.01 & 0.01 \\
Learning rate & $3\times10^{-4}$  & $5\times10^{-4}$  \\
Learning rate schedule & Warmup-then-stable & Warmup-then-stable  \\
Warmup steps & 6000  & 2000 \\
Global batch size & 1024  & 1024 \\
Sequence length & 8192  & 8192 \\
Optimizer & AdamW  & AdamW \\
$\beta_1$ & 0.9 & 0.9  \\
$\beta_2$ & 0.95 & 0.95 \\
Weight decay & 0.1  & 0.1 \\
Clip gradient & 1.0  & 1.0 \\
Adam epsilon & $1\times10^{-15}$ & $1\times10^{-15}$ \\
\bottomrule
\end{tabular}
}
\label{tab:pretrain-stage1-details}
\end{table}

\newpage
\begin{table}[!hbp]
\centering
\caption{Pre-training Stage-2 Details of Loopie-20B-A2B and Loopie-6B-A0.6B}
\vspace{2pt}
{\small
\setlength{\tabcolsep}{20pt}
\renewcommand{\arraystretch}{1.5}
\begin{tabular}{l|cc}
\toprule
 & Loopie-20B-A2B & Loopie-6B-A0.6B  \\
\midrule
Learning rate & $3\times10^{-4}$  & $5\times10^{-4}$  \\
Learning rate schedule & Constant & Constant  \\
MoE Auxiliary loss coeffient & 0.01 & 0.01 \\
Global batch size & 1024  & 1024 \\
Sequence length & 8192  & 8192 \\
Optimizer & AdamW  & AdamW \\
$\beta_1$ & 0.9 & 0.9  \\
$\beta_2$ & 0.95 & 0.95 \\
Weight decay & 0.1  & 0.1 \\
Clip gradient & 1.0  & 1.0 \\
Adam epsilon & $1\times10^{-15}$ & $1\times10^{-15}$ \\
\bottomrule
\end{tabular}
}
\label{tab:pretrain-stage2-details}
\end{table}

\vspace{20mm}
\appsection{Scaling Ladder Details}
\label{app:scaling-ladder}

\begin{table}[!htbp]
    \centering
    \setlength{\tabcolsep}{6pt}
    \renewcommand{\arraystretch}{1.5}
    \scalebox{0.85}{%
    \begin{tabular}{@{}lccccccccccc@{}}
        \toprule
        Rung & Model & \makecell{Total\\Params} & \makecell{Active\\Params} & $L$ & $D$ & $D_{\mathrm{MoE}}$ & Heads & \makecell{Head\\Dim} & $N$ & \makecell{Width/\\Depth} & Tokens \\
        \midrule
        1 & Vanilla & 1.34B & 0.15B & 27 & 640  & 192 & 20 & 32 & 1 & 23.70 & 150B \\
          & Loopie  & 1.08B & 0.11B & 15 & 704  & 256 & 22 & 32 & 2 & 23.47 & 150B \\
        \midrule
        2 & Vanilla & 2.37B & 0.25B & 30 & 768  & 256 & 24 & 32 & 1 & 25.60 & 250B \\
          & Loopie  & 1.81B & 0.18B & 17 & 832  & 320 & 26 & 32 & 2 & 24.47 & 250B \\
        \midrule
        3 & Vanilla & 4.76B & 0.51B & 36 & 1280 & 320 & 32 & 40 & 1 & 35.56 & 500B \\
          & Loopie  & 3.78B & 0.41B & 19 & 1280 & 384 & 32 & 40 & 2 & 33.68 & 500B \\
        \midrule
        4 & Vanilla & 9.14B & 1.00B & 46 & 1280 & 384 & 40 & 32 & 1 & 27.83 & 500B \\
          & Loopie  & 6.36B & 0.68B & 25 & 1408 & 448 & 44 & 32 & 2 & 28.16 & 500B \\
        \bottomrule
    \end{tabular}%
    }
    \caption{Architecture details for the Loopie scaling ladder. Each Loopie model is paired with a non-recurrent vanilla MoE baseline under matched pre-training wall time. Here $D$ denotes hidden dimension, $D_{\mathrm{MoE}}$ denotes each expert's hidden size, $L$ denotes the number of stored Transformer/MoE layers, Heads denotes the number of attention heads, Head Dim denotes the dimension of each attention head, $N$ denotes the number of recurrent layer-loop steps, and Width/Depth denotes $D/(L N)$.}
    \label{tab:loopie-scaling-ladder}
\end{table}

\newpage
\appsection{Supervised Pre-training Details}
\begin{table}[!htbp]
\centering
\caption{Training hyperparameters for Loopie-20B-A2B and Loopie-6B-A0.6B in Supervised Pre-training.}
\vspace{2pt}
{\small
\setlength{\tabcolsep}{20pt}
\renewcommand{\arraystretch}{1.5}
\begin{tabular}{l|cc}
\toprule
 & Loopie-20B-A2B & Loopie-6B-A0.6B  \\
\midrule
Global batch size & 1024  & 1024 \\
Sequence length & 131072 & 131072 \\
Learning rate & $1\times10^{-5}$  & $1\times10^{-5}$  \\
Learning rate schedule & Warmup-then-stable & Warmup-then-stable  \\
Warmup steps & 500 & 500 \\
MoE Auxiliary loss coeffient & 0.01 & 0.01 \\
Optimizer & AdamW  & AdamW \\
$\beta_1$ & 0.9 & 0.9  \\
$\beta_2$ & 0.95 & 0.95 \\
Weight decay & 0.1  & 0.1 \\
Clip gradient & 1.0  & 1.0 \\
Adam epsilon & $1\times10^{-15}$ & $1\times10^{-15}$ \\
\bottomrule
\end{tabular}
}
\label{tab:spt_details}
\end{table}

\end{document}